\pdfoutput=1
\documentclass[10pt,twocolumn]{article}

\usepackage{cvpr}
\usepackage{times}
\usepackage{epsfig}
\usepackage{graphicx}
\usepackage{subcaption}
\usepackage{amsmath}
\usepackage{amssymb}
\usepackage[font={small}]{caption}
\usepackage{multirow}
\usepackage{diagbox}
\usepackage{ifthen}
\usepackage{bm}
\usepackage{pifont}% http://ctan.org/pkg/pifont
\usepackage[hang,flushmargin]{footmisc} % no indent for footnote
\usepackage[pagebackref=true,breaklinks=true,colorlinks,bookmarks=false]{hyperref}
\cvprfinalcopy % *** Uncomment this line for the final submission
\usepackage{cleveref}

\usepackage{microtype}
\usepackage{tikz}
\usepackage{svg}
\usepackage{booktabs}
\usepackage{xurl}
\usepackage{hyperref}
\def\cvprPaperID{59} % *** Enter the CVPR Paper ID here

\begin{document}

%%%%%%%%% TITLE
\title{\vspace{-1cm}TGTM: TinyML-based Global Tone Mapping for HDR Sensors}
\author{
	%\vspace{-0.1cm}
	Peter Todorov \hspace{10pt} Julian Hartig \hspace{10pt} Jan Meyer-Siemon \hspace{10pt} Martin Fiedler \hspace{10pt} Gregor Schewior\\
	\vspace{-0.15cm}
	\small{Dream Chip Technologies GmbH} 
	\vspace{0.1cm}
	\\
	{\tt\footnotesize \{peter.todorov, julian.hartig, jan.meyer-siemon, martin.fiedler, gregor.schewior\}@dreamchip.de}\\
	\vspace{-0.7cm}
}

\maketitle
%\thispagestyle{empty}

%%%%%%%%% ABSTRACT
%\vspace{-0.6cm}
\begin{abstract}
\vspace{-0.2cm}
Advanced driver assistance systems (ADAS) relying on multiple cameras are increasingly prevalent in vehicle technology. Yet, conventional imaging sensors struggle to capture clear images in conditions with intense illumination contrast, such as tunnel exits, due to their limited dynamic range. Introducing high dynamic range (HDR) sensors addresses this issue. However, the process of converting HDR content to a displayable range via tone mapping often leads to inefficient computations, when performed directly on pixel data. In this paper, we focus on HDR image tone mapping using a lightweight neural network applied on image histogram data. Our proposed TinyML-based global tone mapping method, termed as TGTM, operates at 9,000 FLOPS per RGB image of any resolution. Additionally, TGTM offers a generic approach that can be incorporated to any classical tone mapping method. Experimental results demonstrate that TGTM outperforms state-of-the-art methods on real HDR camera images by \textbf{up~to~5.85~dB} higher PSNR with \textbf{orders~of~magnitude} less computations.
\end{abstract}

%\vspace{-0.6cm}

%%%%%%%%% BODY TEXT
\section{Introduction}
\label{introduction}

Dynamic range, defined as the ratio between the brightest and darkest perceivable light levels, is crucial for capturing accurate images. Commonly used image sensors in digital cameras, such as in iPhone 15, have a dynamic range of about 70~dB~\cite{iphone_15_dynamic_range}. However, an intense glare from oncoming car headlights at night or the sun positioned low on the horizon can blind such standard sensors~\cite{hdr_shopovska}. One way to expand dynamic range is to capture and fuse multiple exposures of different lengths into one image~\cite{multiple_exposure_hdr_fusion}. However, this can cause ghosting artifacts for fast moving objects. Consequently, HDR sensors with a dynamic range up to 140~dB have been developed to exactly address the challenging environments of automotive applications. Commonly used cameras in automotive applications typically offer dynamic ranges from 70 to 140~dB. For example, Sony IMX490 sensor~\cite{sony_imx490_sensor}, specifically provides a dynamic range of 140~dB.

In a standard camera setup, a dedicated image signal processor (ISP) is responsible for processing the captured sensor image to produce the final output image. Within the image processing pipeline, there are various steps, one of which is tone mapping. Tone mapping adjusts the intensity of input pixels to improve contrast and enhance color tones in the final image. Usually, including in our approach, the same tone mapping gain is applied to all three color channels (R, G, and B) of the image.

\ifdefined\ifConferenceTemplate\else
    \begin{figure}[t!]
        \centering
        \begin{subfigure}{\linewidth}
            \centering
            \includegraphics[width=\linewidth]{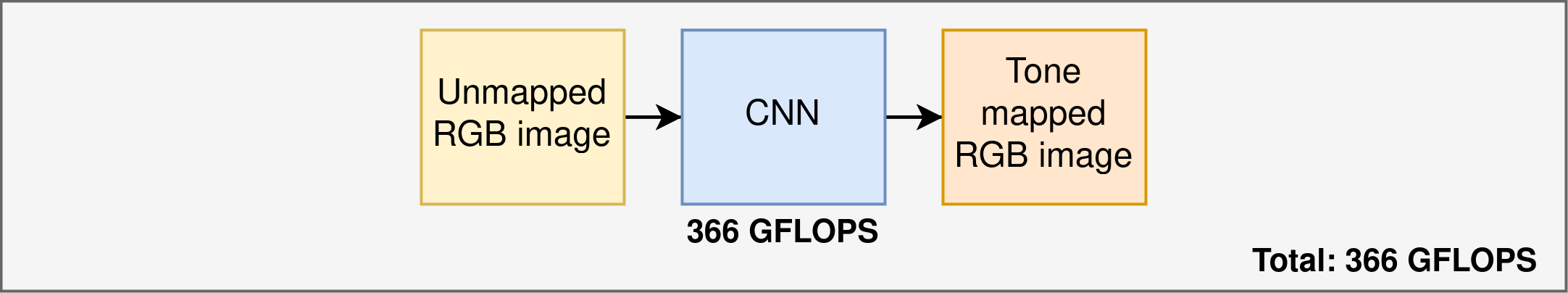}
            \caption{Traditional image-to-image CNN.~\cite{hdr_shopovska}}
        \end{subfigure}
        \par\bigskip
        \begin{subfigure}{\linewidth}
            \centering
            \includegraphics[width=\linewidth]{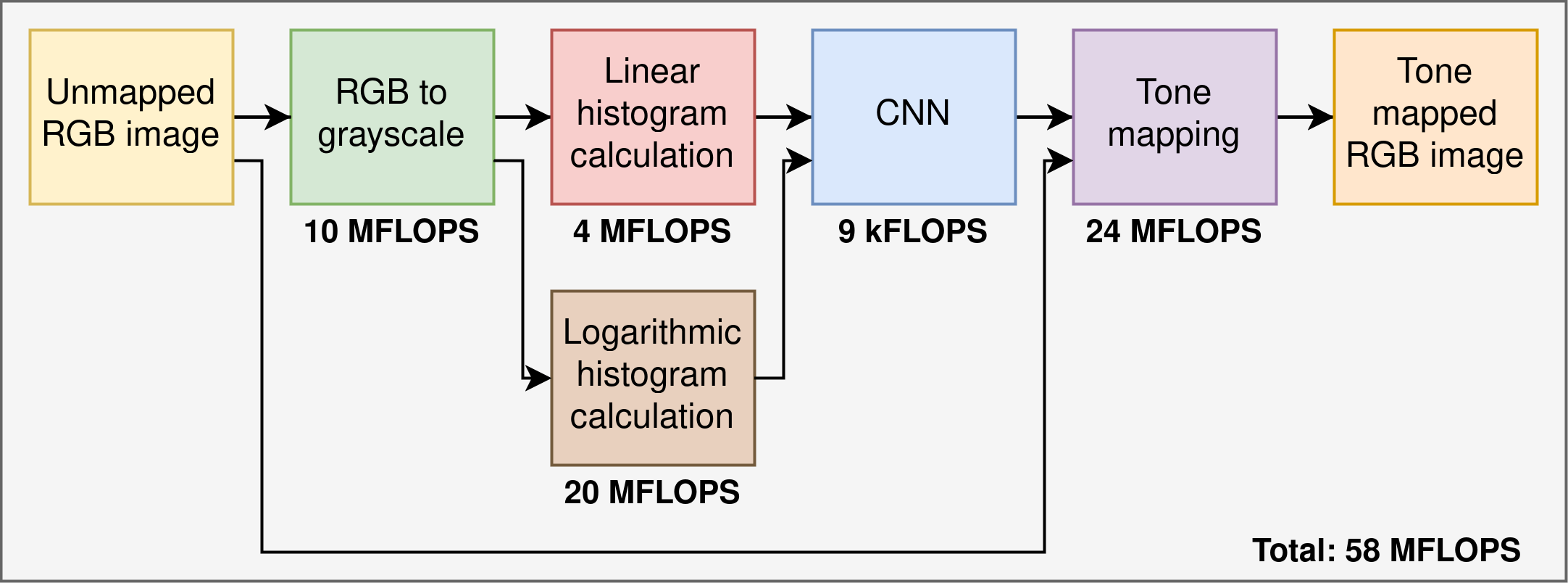}
            \caption{Our proposed CNN-based tone mapping architecture.}
        \end{subfigure}
        \caption{Breakdown of computation in different phases on a 2~Mpx Full HD image.}
        \label{fig:all_ops_breakdown}
    \end{figure}
\fi

Tone mapping can be achieved by either preserving the original dynamic range or by deriving a low dynamic range (LDR) image from an HDR input image. This dynamic range compression~\cite{pizer1987adaptive,zuiderveld1994contrast,reinhard2005dynamic,tmonet,guo2021deepto,tm_using_image_histograms_2} is necessary to ensure compatibility with standard dynamic range (SDR) displays. In this paper, one of our focus areas is to transform a 26-bit HDR image into a 12-bit LDR image.

There are two primary approaches to apply tone mapping. The first involves creating a single tone curve for the entire image, referred to as global tone mapping (GTM)~\cite{he2020conditional,zeng_2020}. The second method entails estimating separate tone curves for different regions of the image, known as local tone mapping (LTM)~\cite{ltmnet}. LTM is particularly effective in preserving local details. However, it can sometimes result in halo artifacts and blocky regions, especially when there are significant differences in tone curves across the image. In contrast, GTM does not encounter these issues, but lacks the ability to enhance local regions as effectively as LTM. Another option to improve tone mapping is to employ pixel-wise enhancement techniques~\cite{ltmnet,hdr_shopovska,hdr_stojkovic,pixel_wise_tm,pixel_wise_tm_2}, often implemented using convolutional neural networks (CNNs). However, these approaches often come with high computational demands compared to methods leveraging image statistics~\cite{tm_using_image_histograms,tm_using_image_histograms_2,tm_using_image_histograms_3}.

In this study, we showcase a computationally efficient tone mapping technique using a CNN to derive tone curve function parameters from image histograms. This type of research, with a particular emphasis on executing machine learning (ML) models on embedded systems, is also referred to as TinyML. The approach, illustrated in Figure~\ref{fig:all_ops_breakdown}, involves applying tone mapping to a 26-bit HDR image and converting it to a 12-bit LDR image. Given that we employ a supervised learning approach to train the machine learning model, we also introduce a suitable data simulation method to transition from a set of 8-bit sRGB images to unmapped linear HDR RGB images. Ultimately, the results with simulated images are compared with available ground truth data, and the results with real 26-bit images captured using an HDR sensor are compared against a manual tone mapping conducted by an image processing engineer within our company.

\section{Related Work}

Recent work on HDR tone mapping designed for automotive applications, DI-TM~\cite{hdr_shopovska}, showcased the simulation of HDR data from SDR data, global tone mapping, and its impact on object detection rates with enhanced image quality. DI-TM compresses 24-bit HDR data into an 8-bit representation to optimize perception, particularly in objects of interest. Moreover, DI-TM addresses the scarcity of HDR training data by simulating it. During data simulation, DI-TM adjusts noise, contrast, and color temperature, indicating changes in these aspects during tone mapping. The enhanced version of DI-TM, to which we refer as DI-DMTM~\cite{hdr_stojkovic}, further considers robustness to demosaicing artifacts. These methods contrast with our approach, where we solely modify the global tone of the image. Although DI-TM offers a functional solution, its reliance on pixel data as input leads to high computational load, compromising its efficiency. The model consists of 340,227 trainable parameters, with 366 billion floating-point operations (FLOPS) required for images at a resolution of 1920x1080 pixels.

Well-known classical state-of-the-art methods in tone mapping include Reinhard~\cite{reinhard} and Farbman~\cite{farbman_2008_edge}. The Reinhard algorithm is noted for its computational efficiency and consistent outcomes, especially when compared to neural network-based approaches. It excels at preserving global contrast and details. However, to achieve satisfactory results across different images and lighting conditions, it may require manual parameter adjustments by the user. In contrast, Farbman's approach operates on multiple scales, decomposing the image into different frequency bands and treating each band separately before recombining them to form the final tone-mapped image. While Farbman's approach offers advantages in terms of local detail preservation and halo reduction, it comes with a drawback in increased computational complexity. The Farbman algorithm requires processing multiple scale layers and performing complex recombination steps. This makes Farbman's method less suitable for real-time or resource-constrained applications. Additionally, fine-tuning the parameters of the algorithm to achieve optimal results may require experimentation, as the effectiveness of the tone mapping process can vary depending on the input image. Nevertheless, both methods can deliver high-quality results, but they often require manual parameter adjustments tailored to each specific scene, thereby lacking full automation.

Embedded ISP pixel streaming pipelines often impose strict memory and latency constraints. For instance, techniques relying on a frame buffer, such as the multi-scale base-detail decomposition in Farbman and the image-to-image CNN approaches in DI-TM and DI-DMTM, are usually impractical. Consequently, solutions like the Reinhard method fulfill these requirements. Thus, drawing inspiration from Reinhard, in our proposed solution we aim to preserve the advantages of Reinhard while addressing the challenge of automatic parameter tuning.

\section{Proposed Solution}
\label{sec:proposed_solution}

\begin{figure*}
    \centerline{\includegraphics[width=\textwidth]{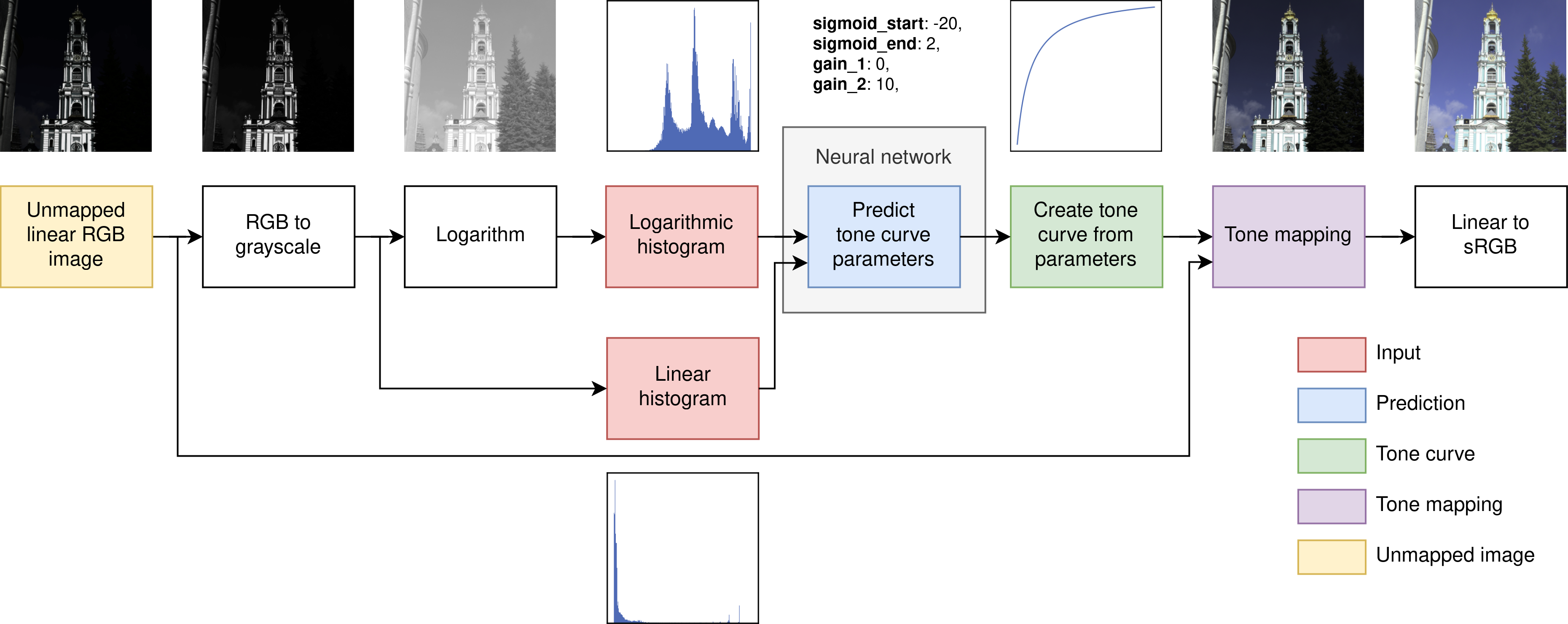}}
    \caption{Inference pipeline. The CNN uses two image histograms to output four parameters for tone curve creation.}
    \label{fig:inference_pipeline}
\end{figure*}

In this work, we propose a computationally efficient global tone mapping model designed to run in a modern ISP. This study has been developed in the context of the research project ZuSE-KI-Mobil~- "KI-Akzelerator SoC Design" \cite{zukimo_1,zukimo_2} funded by the German government (BMBF) under the grant number 16ME0985. Part of the goal of this project is to improve Dream Chip's automotive image processing pipeline IP (RPP) using CNN algorithms without impacting or with minimum impact to the processing latency. Furthermore, it should be demonstrated how this ISP CNN method can be implemented on the ZuSE-KI-Mobil SoC with or without SoC design modifications. Given the efficiency of the CNN method, requiring only 9k FLOPS, the algorithm can run on the ZuSE-KI-Mobil SoC's application processing unit (APU), eliminating the need for a dedicated CNN accelerator.

To avoid the burden of collecting and manually tone mapping a training dataset, we introduce a data simulation method for a supervised learning approach. This includes corresponding pairs of input images and ground truth tone curves. There are four key ideas in the proposed solution:
\begin{enumerate}
    \itemsep 0.1em
    \item \textit{Data simulation with inverted tone mapping curves.} Inverting a tone mapping curve and applying it to an image automates the generation of a custom dataset.
    \item \textit{Data compression with image histograms.} Utilizing a combination of linear and logarithmic histograms for data compression, enabling the approach to handle information loss while maintaining resolution independence.
    \item \textit{Prediction of curve parameters.} Instead of predicting discrete tone curve values, the approach predicts curve parameters that are passed to an analytical function for computing the curve. This ensures functional safety by leveraging known limits and behaviors.
    \item \textit{Curve integral loss.} Introducing a curve integral loss to mitigate bias in the error, ensuring higher accuracy.
\end{enumerate}

\subsection{Inference Pipeline}

The complete pipeline for tone mapping an image is illustrated in Figure~\ref{fig:inference_pipeline}. The pipeline operates on an unmapped linear RGB HDR image as input with a dynamic range of $[0, 2^{bits} -1]$. First, the image is converted to luminance space with corresponding weights assigned to the R, G, and B components [0.25, 0.50, 0.25]. Subsequently, the linear histogram is computed from this transformed image. In a parallel path, the logarithmic histogram is calculated. Before the logarithm operation, the image needs to be converted to the same bit-width used during the training data simulation. This bit-width conversion is important as logarithm is a nonlinear operation. However, in our case, the conversion is unnecessary as both pipelines use 26-bit format. Both the linear and logarithmic histograms are then fed into the neural network, which predicts four parameters, see Table~\ref{tab:reinhard_parameters}, for tone curve creation.

The tone curve is created by using a modified Reinhard equation
\begin{equation}
    c_i(x) = \frac{x \cdot m \cdot (1 + g_i)}{g_i \cdot x + m}
    \label{eq:reinhard}
\end{equation}
where $x$ is the input value between $0$ and $m$, $m$ is the maximum value used for the curve, for example $2^{bits} - 1$, and $g_i$ is a gain. The equation is extended from Reinhard~\cite{reinhard} by a scaling factor so that white is preserved as white, in other words, $m$ maps to $m$. The same is for 0 ensuring $0 \leq c_i \leq m$. The final tone curve is created by blending two of these curves, $c_1$ and $c_2$, which enhances tone mapping in case of having bright and dark parts in the image. The blending is done with a weighting from the sigmoid curve $s(s_s, s_e)$ where $s_s$ is the start value and $s_e$ is the end value of sigmoid. The tone curve looks as follows
\begin{equation}
    r(x) = s(x, s_s, s_e) \cdot c_1(x) + (1 - s(x, s_s, s_e)) \cdot c_2(x)
    \label{eq:modified_reinhard}
\end{equation}
where $s$ is the value of sigmoid in the point $x$, and $c_1$ and $c_2$ are the gain curves evaluated in the point $x$ with respective gains $g_1$ and $g_2$. In case a negative slope is created by the blending, it will be replaced with a horizontal line starting from the top point and ending to a point higher than the horizontal line. This fixes the unnatural outcome which would otherwise occur due to inverting parts of the tonal range. Finally, the tone mapping, which includes the conversion from 26-bit to 12-bit, is applied to the unmapped image followed by a conversion to sRGB domain.

\subsection{Data Simulation}
\label{sec:data_simulation}

\begin{figure*}
    \centerline{\includegraphics[width=\textwidth]{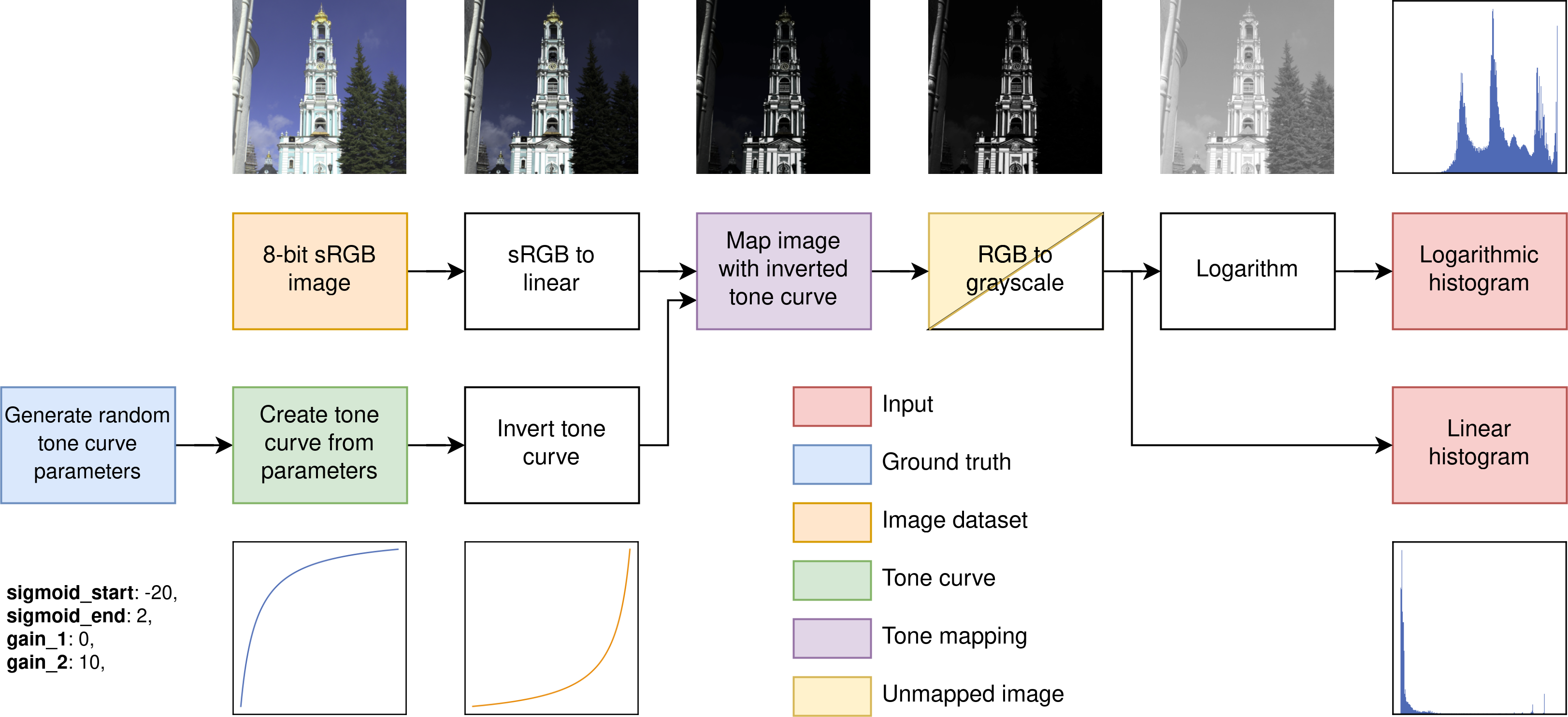}}
    \caption{Data simulation pipeline. An unmapped image is created by applying an inverted tone curve (orange) to a ground truth image.}
    \label{fig:data_simulation_pipeline}
\end{figure*}

A robust tone mapping dataset should avoid relying on one individual's preferences for tone mapping. Additionally, tone mapping accuracy should not be limited by the use of existing tone mapping algorithms to generate ground truth data. We evaluated two public datasets: HDR+ dataset~\cite{hasinoff2016burst} (Creative Commons license: CC-BY-SA) and MIT-Adobe FiveK~\cite{fivek} (Research license). HDR+ dataset provides the meta data for each image, including tone mapping curves used in processing the raw images. However, this is not sufficient as we do not want to limit our capabilities and in the best case reach the level of an already existing tone mapping algorithm. MIT-Adobe FiveK, on the other hand, has employed five photography students in an art school to adjust the tone of the photos. While this approach provides human input, it is not ideal as local adjustments were made to the images, making it challenging to accurately estimate a single tone curve between two images. Therefore, we propose to simulate a dataset without relying on existing algorithms or without the need of human labor in the tuning process.

A typical data simulation approach in image enhancement projects is to create an input image by adding degradation to a high quality ground truth image. In Figure~\ref{fig:data_simulation_pipeline}, we pick an 8-bit sRGB image from MIT-Adobe FiveK dataset, convert it to 26-bit format to reduce quantization errors in the subsequent processing steps, and apply an inverse gamma curve to get linear RGB data. Meanwhile in a separate branch, a random tone curve is generated using Equation~\ref{eq:modified_reinhard}. The four parameters, used in the curve creation, are randomly picked with a uniform distribution over the defined range except the gain 2 parameter. The parameters are in Table~\ref{tab:reinhard_parameters}. The gain 2 parameter exhibits non-uniform randomness due to its nonlinear relationship with the curve shape. Using uniform randomness would result in an average gain value of around 10k, predominantly consisting images from extremely dark scenes. Therefore, to ensure equal representation of bright and dark scenes, we apply randomness in the logarithmic domain. The impact of a gain is illustrated in Figure~\ref{fig:applying_inverted_curve}.

\begin{table}[t]
    \caption{The four Reinhard curve parameters with their value ranges and random distribution used in the data simulation.}
    \label{tab:reinhard_parameters}
    \vskip 0.15in
    \begin{center}
    \begin{small}
    \begin{rmfamily}
    \ifdefined\ifConferenceTemplate
        \begin{tabular}{lccc}
    \else
        \begin{tabular}{l|c|c|c}
    \fi
    \toprule
    Parameter & Notation & Range & Distribution \\
    \midrule
    Sigmoid start & $s_s$ & [-20, 2] & Uniform \\
    Sigmoid end   & $s_e$ & [2, 20] & Uniform \\
    Gain 1        & $g_1$ & [0, 3] & Uniform \\
    Gain 2        & $g_2$ & [3, 20000] & Non-uniform \\
    \bottomrule
    \end{tabular}
    \end{rmfamily}
    \end{small}
    \end{center}
    \vskip -0.1in
\end{table}

The tone curve is created according to Equation~\ref{eq:modified_reinhard} and then it is inverted. Subsequently, the image is tone mapped with the inverted tone curve, followed by transformation to luminance space. Lastly, the two histograms are computed from the luminance image.

Given the wide dynamic range involved, most of the pixel values fall within visually dark regions. To demonstrate the effect of a tone curve gain, see Figure~\ref{fig:applying_inverted_curve}. It illustrates the full dynamic range used during data simulation. Additionally, it highlights the need for logarithmic histograms to avoid the information loss in linear histograms.

\subsection{Neural Network Architecture}

The neural network architecture is depicted in Figure~\ref{fig:model_architecture}. The input is a stacked tensor of 2x256-bin histograms. This arrangement ensures alignment of histograms with corresponding elements during convolution. The histograms are normalized to ensure consistent performance regardless of image resolution.

Computing mean average error (MAE) from the four parameters in Table~\ref{tab:reinhard_parameters} is effective, but relying solely on it will introduce bias in the loss. Given the strong nonlinear relationship between the value of parameter $g_2$ and the shape of the respective gain curve $c_2$, it is recommended to compute the loss also from $c_2$, using Equation~\ref{eq:reinhard}, to mitigate bias. For efficiency, we propose computing this curve loss between the integrals of predicted and ground truth $c_2$ gain curves calling it \textit{curve integral loss} (CIL).  So, during the training, the cost function comprises a combination of MAE and CIL.

\newlength{\figurewidth}
\setlength{\figurewidth}{0.6\columnwidth}
\ifdefined\ifConferenceTemplate\else
\begin{figure*}
    \centering
    % 1st row
    \begin{subfigure}{0.181\linewidth}
        \begin{tikzpicture}
            \node[anchor=north west,inner sep=0] (image) at (0,0) 
            {\includegraphics[width=\linewidth]{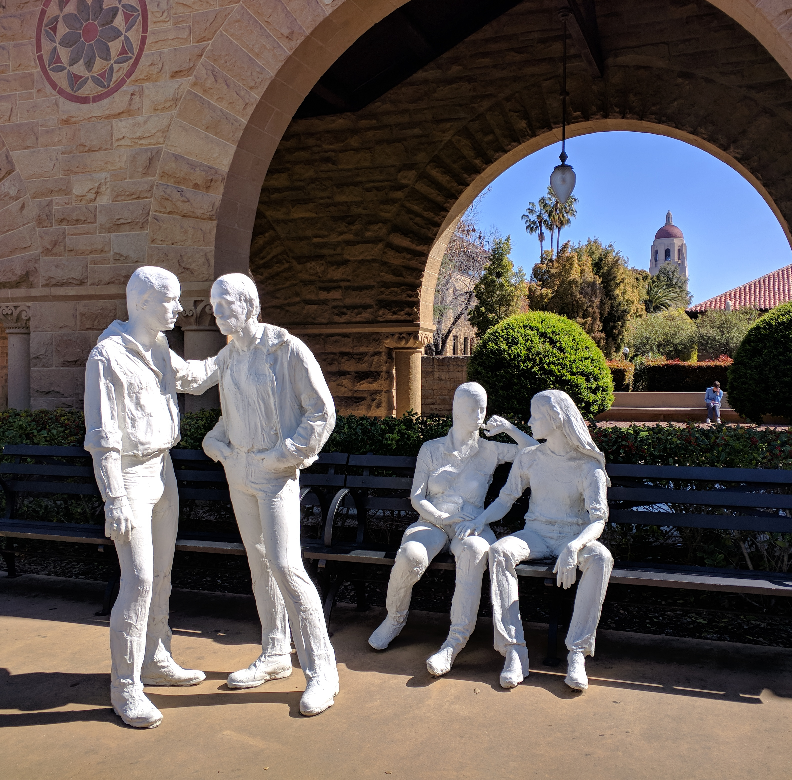}};
            \node[scale=0.6,fill=white,fill opacity=0.8,anchor=north west,text width=\linewidth-30pt,
            inner xsep=0pt,inner ysep=4pt,outer ysep=0.3pt,align=center,text opacity=1,font=\LARGE] at ([xshift=0pt]image.north west){Source};
            % from https://tex.stackexchange.com/a/9562/121799
        \end{tikzpicture}
    \end{subfigure}
    \begin{subfigure}{0.181\linewidth}
        \includegraphics[width=\linewidth]{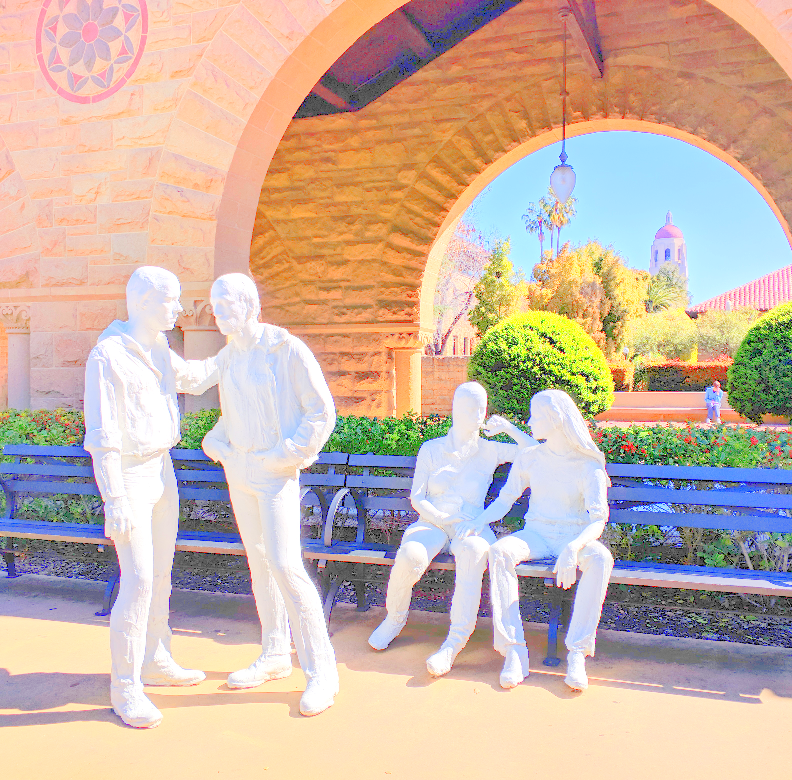}
    \end{subfigure}
    \begin{subfigure}{0.181\linewidth}
        \includegraphics[width=\linewidth]{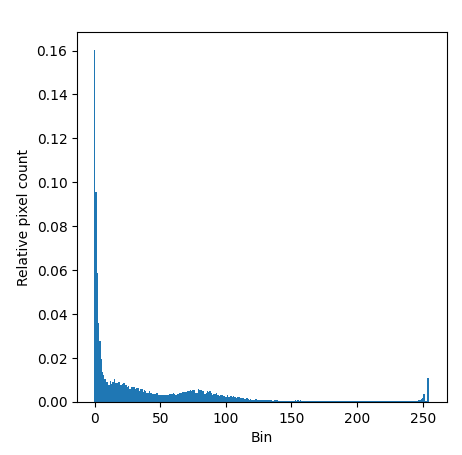}
    \end{subfigure}
    \begin{subfigure}{0.181\linewidth}
        \includegraphics[width=\linewidth]{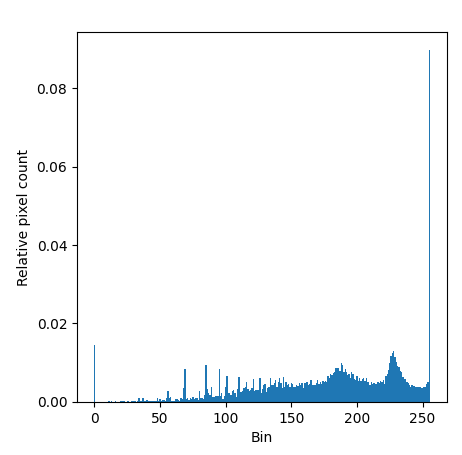}
    \end{subfigure}
    \begin{subfigure}{0.181\linewidth}
        \includegraphics[width=\linewidth]{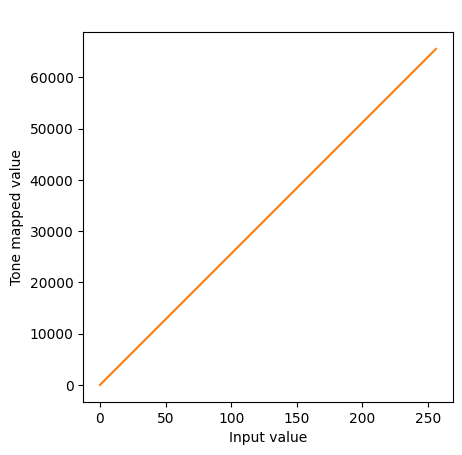}
    \end{subfigure}
    \begin{subfigure}{0.04\linewidth}
        \centering
        \raisebox{0.56in}{\rotatebox[origin=c]{90}{0}}  %% \raisebox is needed in arXiv
    \end{subfigure}
    \vspace{0.12cm}
    \linebreak
    %\par  % Commented in arXiv
    % 2nd row
    \begin{subfigure}{0.181\linewidth}
        \includegraphics[width=\linewidth]{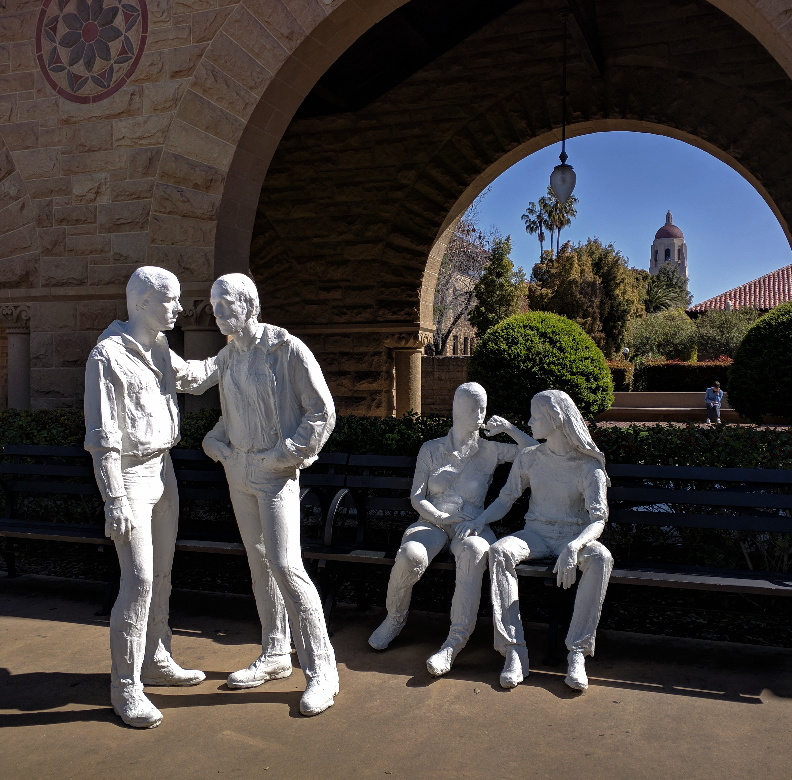}
    \end{subfigure}
    \begin{subfigure}{0.181\linewidth}
        \includegraphics[width=\linewidth]{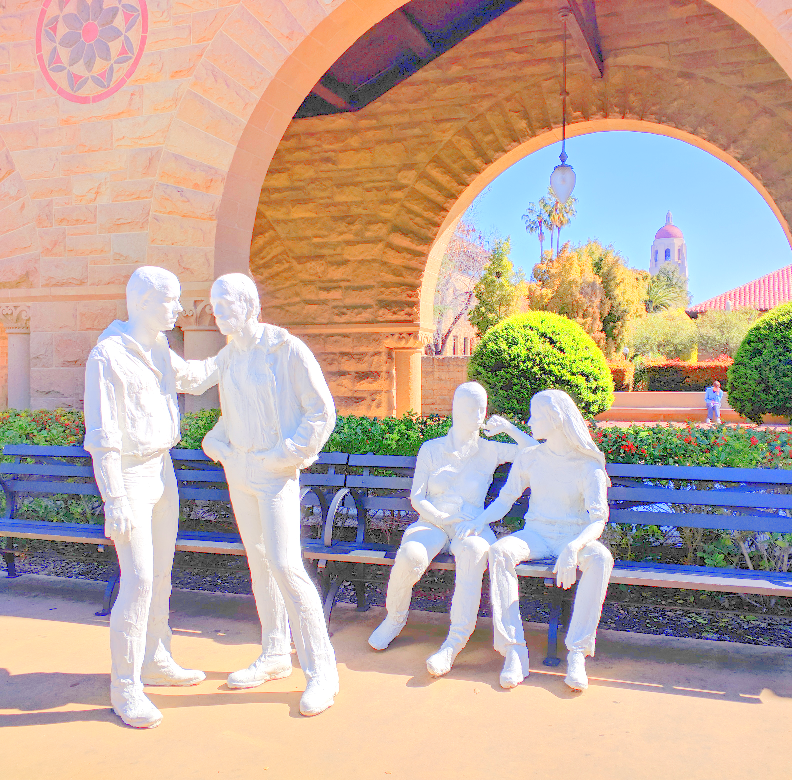}
    \end{subfigure}
    \begin{subfigure}{0.181\linewidth}
        \includegraphics[width=\linewidth]{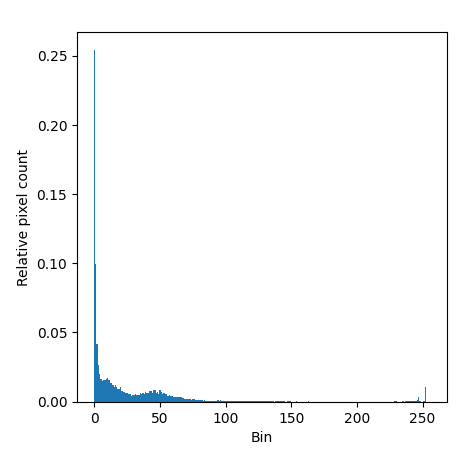}
    \end{subfigure}
    \begin{subfigure}{0.181\linewidth}
        \includegraphics[width=\linewidth]{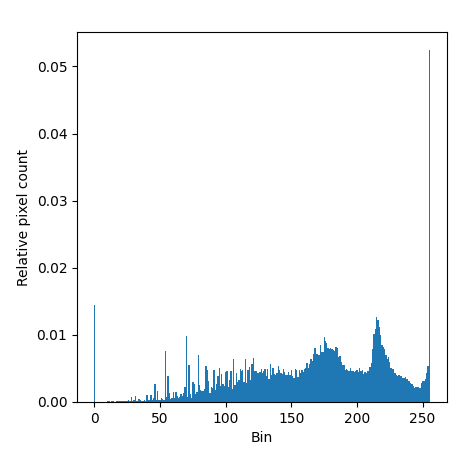}
    \end{subfigure}
    \begin{subfigure}{0.181\linewidth}
        \includegraphics[width=\linewidth]{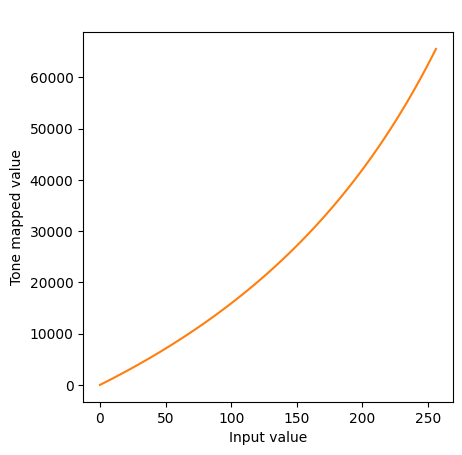}
    \end{subfigure}
    \begin{subfigure}{0.04\linewidth}
        \centering
        \raisebox{0.56in}{\rotatebox[origin=c]{90}{1}}  %% \raisebox is needed in arXiv
    \end{subfigure}
    \vspace{0.06cm}
    \linebreak
    % 3rd row
    \begin{subfigure}{0.181\linewidth}
        \includegraphics[width=\linewidth]{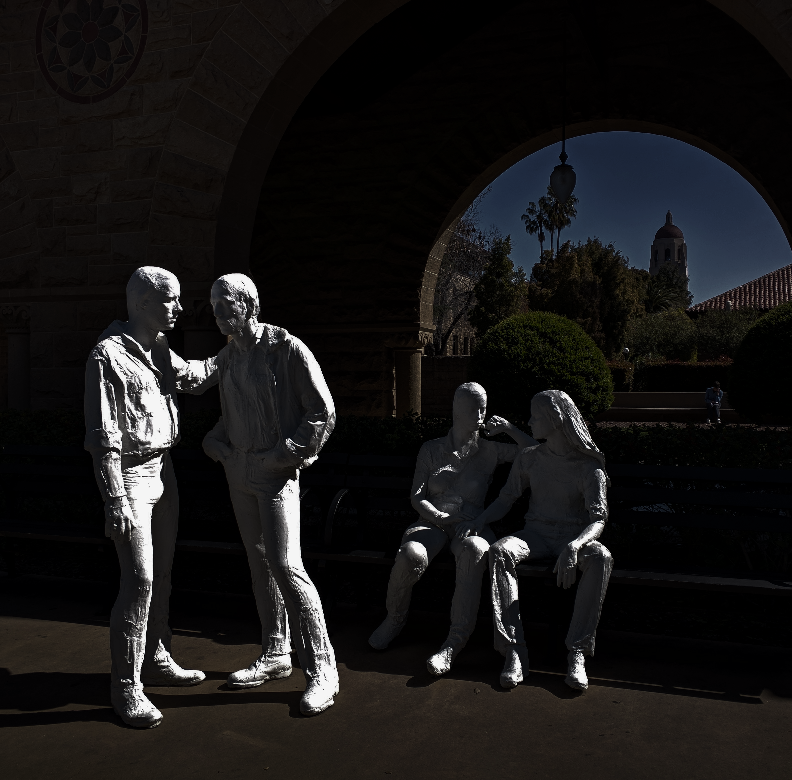}
    \end{subfigure}
    \begin{subfigure}{0.181\linewidth}
        \includegraphics[width=\linewidth]{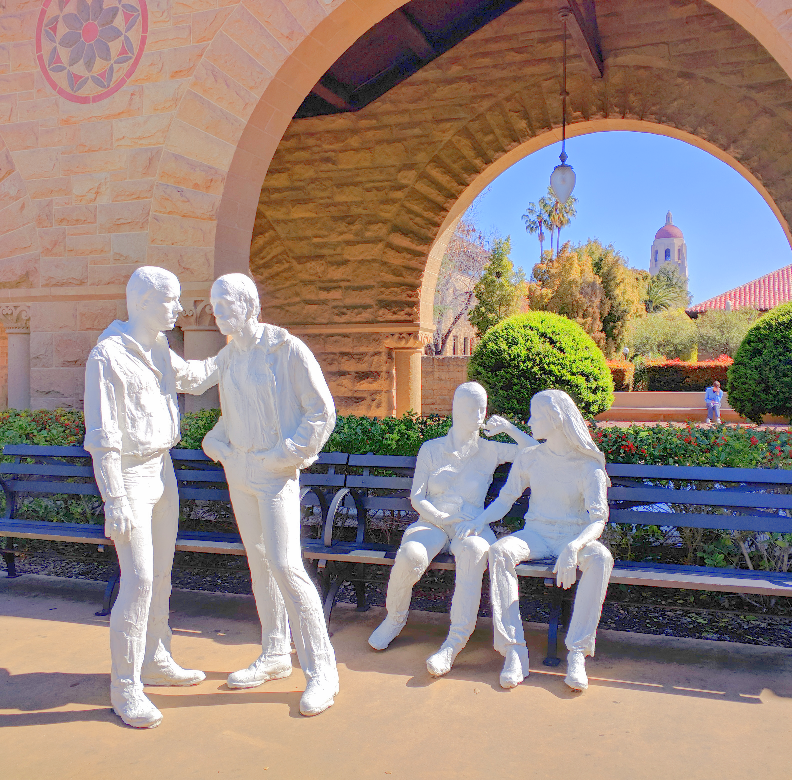}
    \end{subfigure}
    \begin{subfigure}{0.181\linewidth}
        \includegraphics[width=\linewidth]{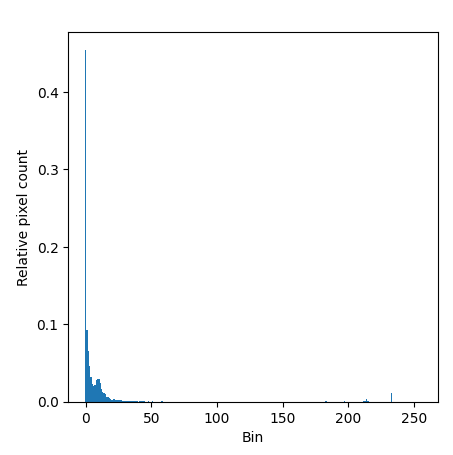}
    \end{subfigure}
    \begin{subfigure}{0.181\linewidth}
        \includegraphics[width=\linewidth]{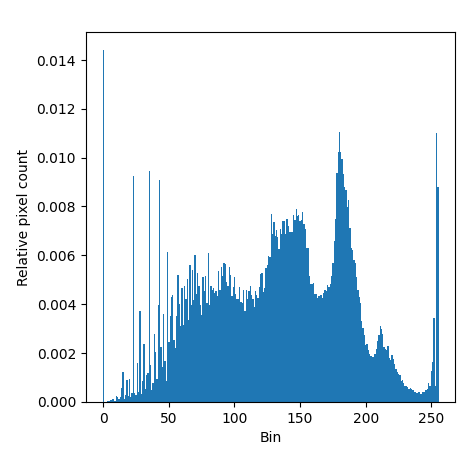}
    \end{subfigure}
    \begin{subfigure}{0.181\linewidth}
        \includegraphics[width=\linewidth]{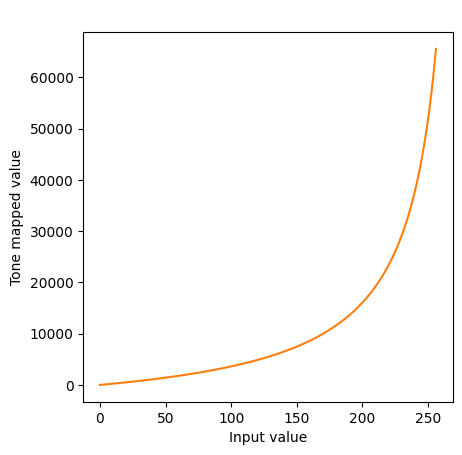}
    \end{subfigure}
    \begin{subfigure}{0.04\linewidth}
        \centering
        \raisebox{0.56in}{\rotatebox[origin=c]{90}{10}}  %% \raisebox is needed in arXiv
    \end{subfigure}
    \vspace{0.06cm}
    \linebreak
    % 4th row
    \begin{subfigure}{0.181\linewidth}
        \includegraphics[width=\linewidth]{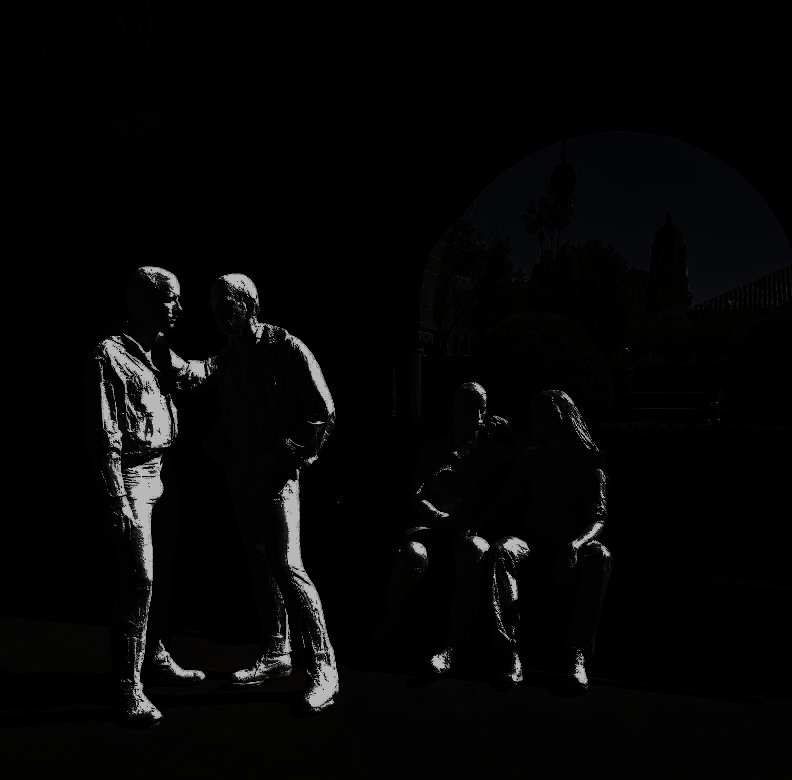}
    \end{subfigure}
    \begin{subfigure}{0.181\linewidth}
        \includegraphics[width=\linewidth]{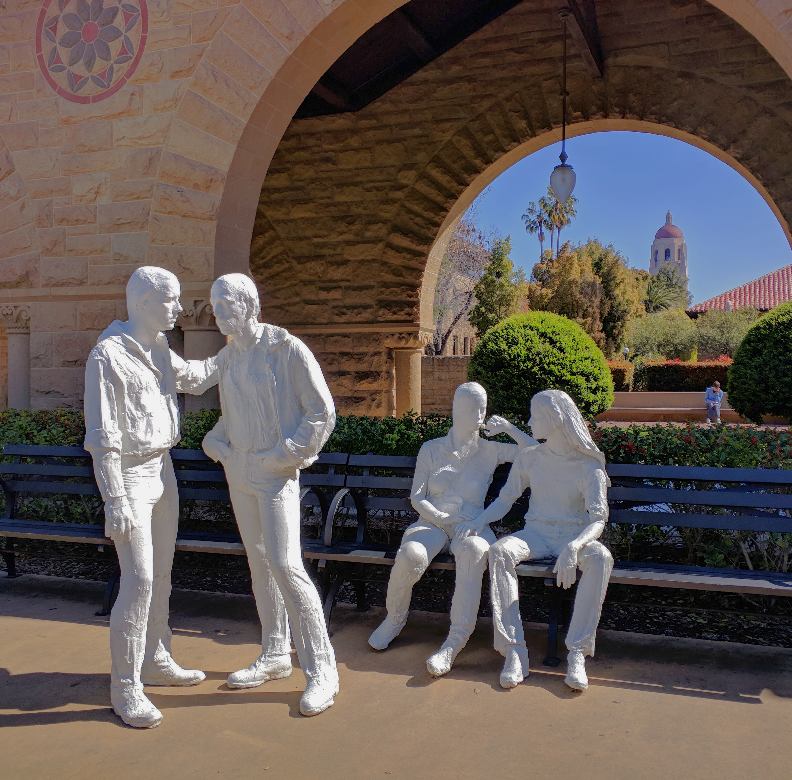}
    \end{subfigure}
    \begin{subfigure}{0.181\linewidth}
        \includegraphics[width=\linewidth]{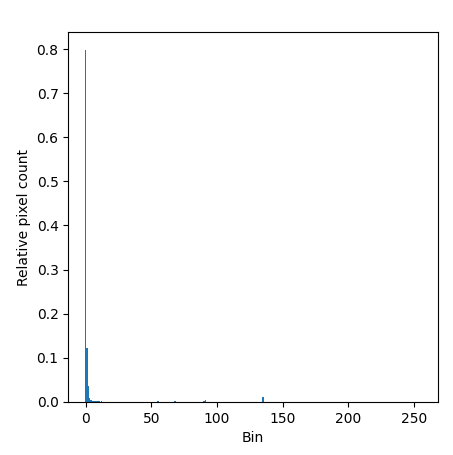}
    \end{subfigure}
    \begin{subfigure}{0.181\linewidth}
        \includegraphics[width=\linewidth]{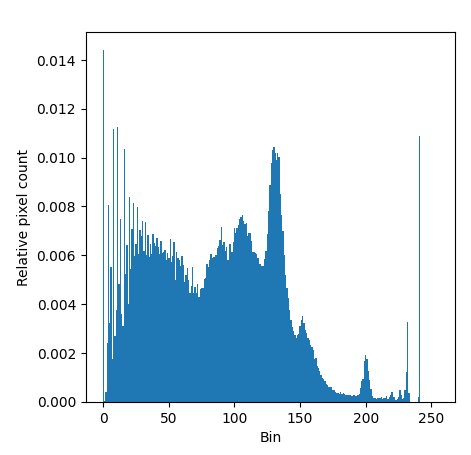}
    \end{subfigure}
    \begin{subfigure}{0.181\linewidth}
        \includegraphics[width=\linewidth]{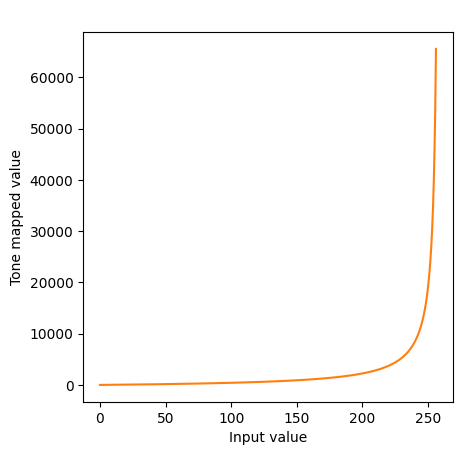}
    \end{subfigure}
    \begin{subfigure}{0.04\linewidth}
        \centering
        \raisebox{0.56in}{\rotatebox[origin=c]{90}{100}}  %% \raisebox is needed in arXiv
    \end{subfigure}
    \vspace{0.06cm}
    \linebreak
    % 5th row
    \begin{subfigure}{0.181\linewidth}
        \includegraphics[width=\linewidth]{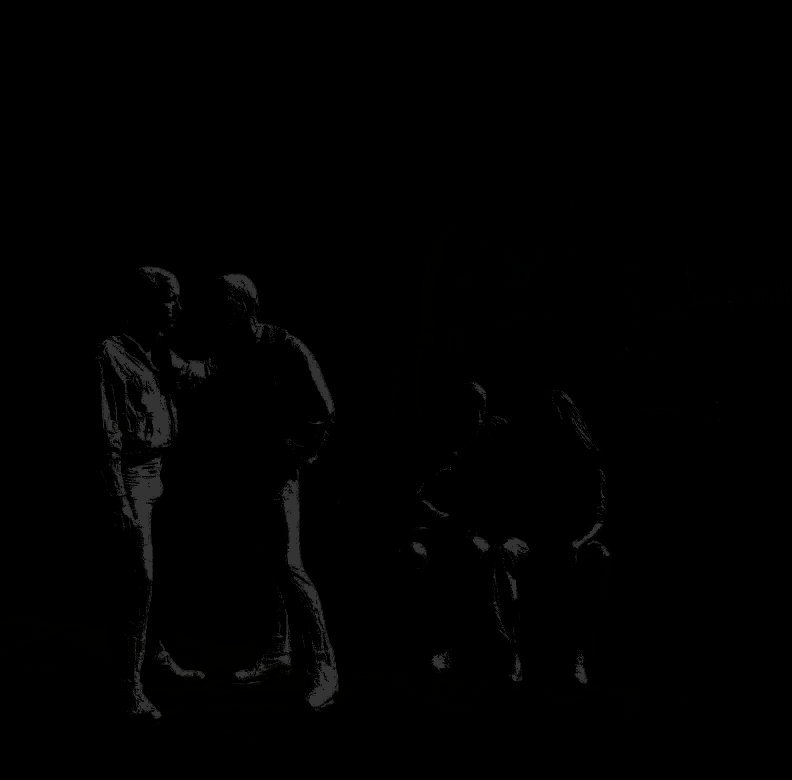}
    \end{subfigure}
    \begin{subfigure}{0.181\linewidth}
        \includegraphics[width=\linewidth]{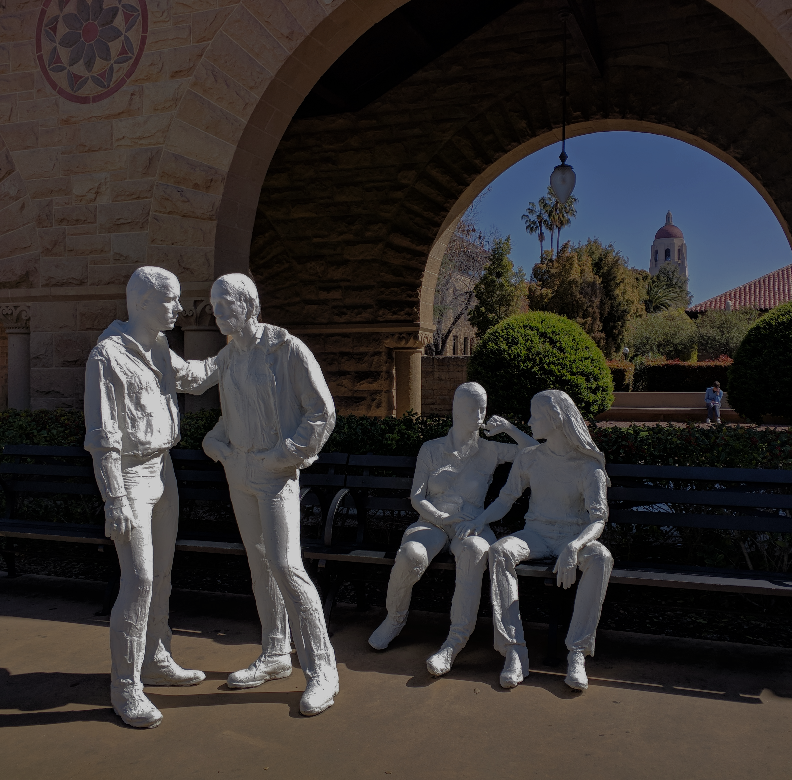}
    \end{subfigure}
    \begin{subfigure}{0.181\linewidth}
        \includegraphics[width=\linewidth]{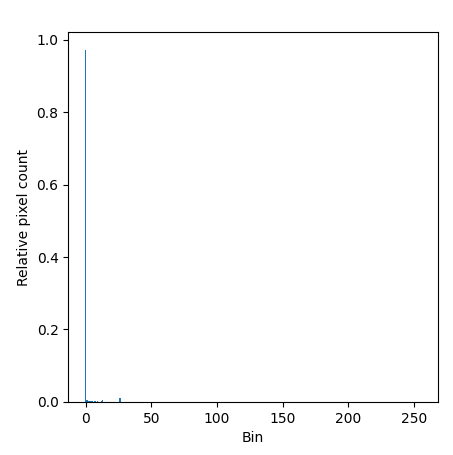}
    \end{subfigure}
    \begin{subfigure}{0.181\linewidth}
        \includegraphics[width=\linewidth]{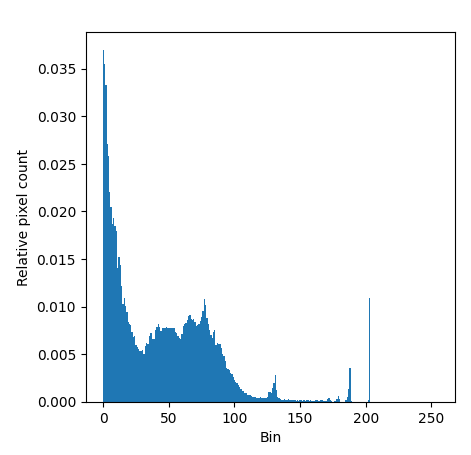}
    \end{subfigure}
    \begin{subfigure}{0.181\linewidth}
        \includegraphics[width=\linewidth]{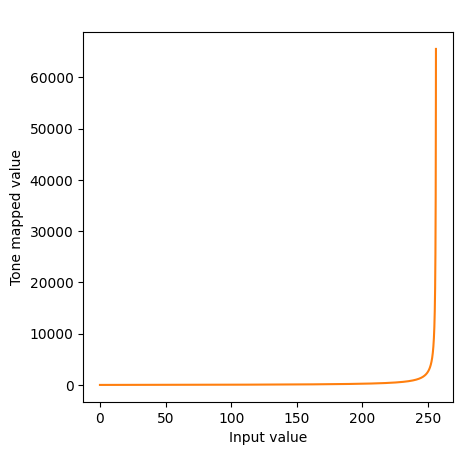}
    \end{subfigure}
    \begin{subfigure}{0.04\linewidth}
        \centering
        \raisebox{0.56in}{\rotatebox[origin=c]{90}{1000}}  %% \raisebox is needed in arXiv
    \end{subfigure}
    \vspace{0.06cm}
    \linebreak
    % 6th row
    \begin{subfigure}{0.181\linewidth}
        \includegraphics[width=\linewidth]{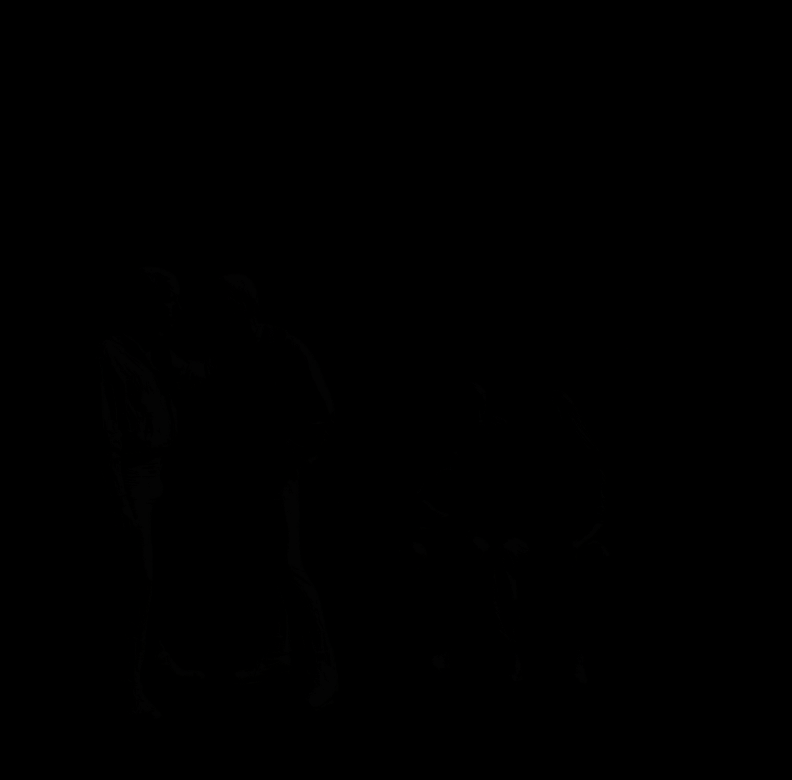}
    \end{subfigure}
    \begin{subfigure}{0.181\linewidth}
        \includegraphics[width=\linewidth]{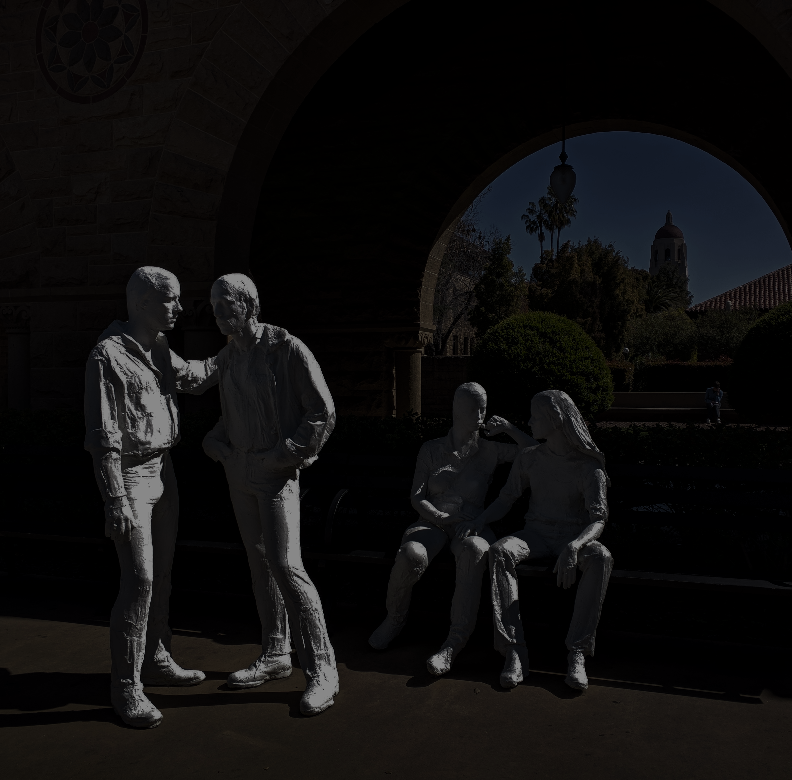}
    \end{subfigure}
    \begin{subfigure}{0.181\linewidth}
        \includegraphics[width=\linewidth]{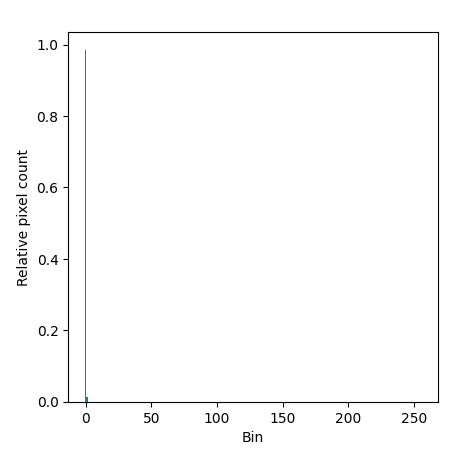}
    \end{subfigure}
    \begin{subfigure}{0.181\linewidth}
        \includegraphics[width=\linewidth]{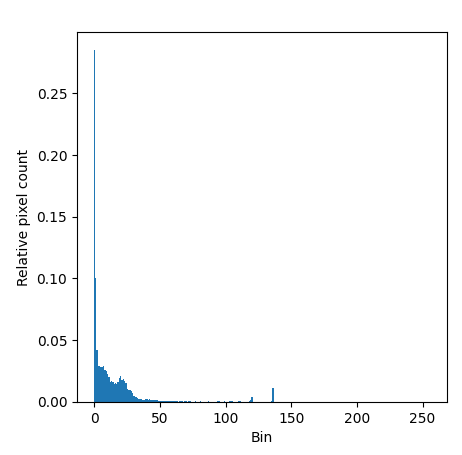}
    \end{subfigure}
    \begin{subfigure}{0.181\linewidth}
        \includegraphics[width=\linewidth]{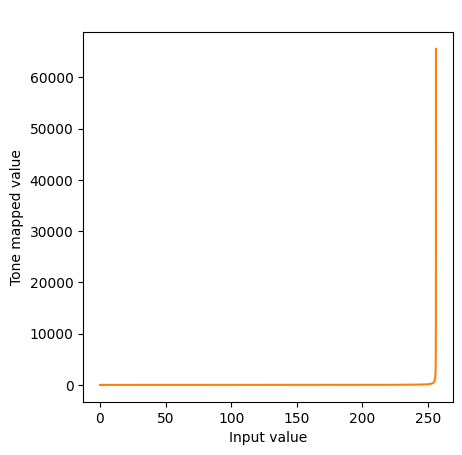}
    \end{subfigure}
    \begin{subfigure}{0.04\linewidth}
        \centering
        \raisebox{0.56in}{\rotatebox[origin=c]{90}{20000}}  %% \raisebox is needed in arXiv
    \end{subfigure}
    \vspace{0.06cm}
    \linebreak
    %\par  % Comment in arXiv
    \begin{subfigure}{0.181\linewidth}
        \caption*{sRGB image}
    \end{subfigure}
    \begin{subfigure}{0.181\linewidth}
        \caption*{Logarithmic image}
    \end{subfigure}
    \begin{subfigure}{0.181\linewidth}
        \caption*{Linear histogram}
    \end{subfigure}
    \begin{subfigure}{0.181\linewidth}
        \caption*{Logarithmic histogram}
    \end{subfigure}
    \begin{subfigure}{0.181\linewidth}
        \caption*{Inverted tone curve}
    \end{subfigure}
    \begin{subfigure}{0.04\linewidth}
        \caption*{Gain}
    \end{subfigure}
    \vspace{-0.1cm}
    \caption{Visual effect of applying an inverted tone curve on the right to the source image at the top left. The higher the curve gain is, the stronger the "L-shape" of the curve is. Whenever the gain is above 25, an image is visually completely black and a linear histogram is ineffective to represent the image information.}
    \label{fig:applying_inverted_curve}
\end{figure*}
\setlength{\figurewidth}{\columnwidth}
\fi

\begin{figure}
    \centerline{\includegraphics[width=\figurewidth]{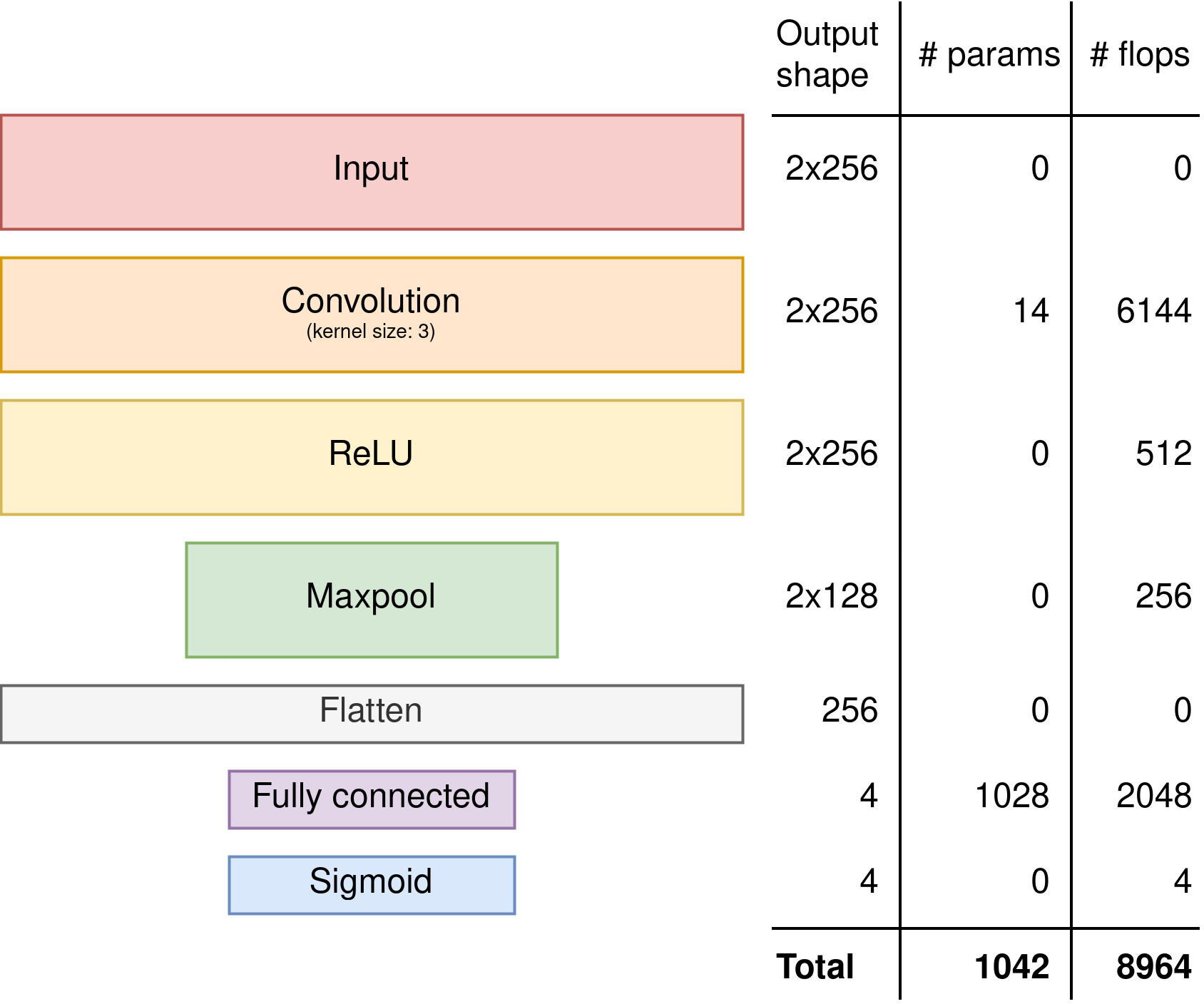}}
    \caption{TGTM model architecture with a breakdown of the number of parameters and FLOPS.}
    \label{fig:model_architecture}
\end{figure}

The predicted output includes four values which are used in Equation~\ref{eq:modified_reinhard} for the creation of the tone curve. This approach worked out better than estimating discrete curve values. Furthermore, by predicting the curve parameters, we ensure consistency regarding all possible curve shapes, smooth transitions across frames, and the ability to generate the curve with any number of steps, even with unequal step sizes. The curve is consistently generated within a fixed range from 0 to $m$. Alternatively, generating the curve from the minimum to maximum captured value of the image may result in better single-image tone mapping. However, it alters the image value range, therefore making the algorithm more vulnerable for temporal inconsistencies, such as flickering lights. For video capturing use cases, we recommend generating the curve for the entire value range to achieve better temporal stability.

\subsection{Computational Complexity}

The proposed network is compact, comprising only 1,042 parameters and requiring 8,964 FLOPS. Our CNN requires only 0.3 \% of the model parameters used in DI-TM (340,227~parameters and 366~GFLOPS). However, when comparing FLOPS, it is essential to consider the entire pipeline of TGTM. Given the need for efficient memory and computation in applying tone mapping, it is impractical to create a tone curve for the entire value range of [0,~$2^{26}-1$], which includes 67 million values. Therefore, we have chosen to create a non-equidistant piecewise linear tone curve with 49 sample points and to interpolate the remaining values during tone mapping.

We have assessed that calculating the 49 tone curve sample points using Equation~\ref{eq:modified_reinhard} costs less than 3,000 FLOPS in total. As this calculation is performed only once per image, alongside the CNN, its impact is insignificant compared to the FLOPS applied to the pixel data, effectively approximating to 0~FLOPS per pixel. For reference, the runtime of the exported C code for the CNN model and tone curve creation was $994~\mu s$ on Raspberry Pi~1~\cite{raspberry_pi_1} and $53~\mu s$ on Raspberry Pi~4~\cite{raspberry_pi_4}. However, for the proposed tone mapping the cost of applying the tone curve to a pixel is 12 FLOPS. Additionally, as shown in Figure~\ref{fig:all_ops_breakdown}, transforming an RGB image to grayscale and computing linear and logarithmic histograms each costs 5, 2, and 10~FLOPS per pixel, respectively. While DI-TM requires 176,614~FLOPS per pixel, our overall cost is substantially lower, totaling 29~FLOPS per pixel. Eventually, a few thousands of FLOPS in our CNN and in the tone curve creation are negligible compared to the parts of operating with pixel data, making the proposed CNN approach computationally implementable to any ISP.

\subsection{Training Details}
\label{sec:training_details}

The training dataset was MIT-Adobe FiveK~\cite{fivek}. It contains 5,000 images with varying resolutions, orientations, and image contents. The training data was simulated with six random inverted tone curves per image leading to 30,000 training samples. The validation data was simulated with a single random inverted tone curve per image applied to 2,000 images. In the simulation, following the pipeline in Figure~\ref{fig:data_simulation_pipeline}, 2x256 input histograms and four ground truth parameters were saved in a text file. It was completed in 15 minutes on a single thread of an AMD Ryzen Threadripper PRO 3975WX CPU.

The used batch size was 16, which led to 1,875 iterations per epoch. The optimizer was Adam~\cite{adam_optimizer} with a learning rate of 1e-4. The used scheduler, ReduceLROnPlateau, was set to reduce learning rate based on validation loss on plateau by a factor of 0.3 until to a minimum of 1e-8, with a patience of 6 epochs. The final model was reached after 97 epochs (181,875 iterations). The development was done using the Pytorch~\cite{pytorch} framework with a total training time of 7 minutes on a single Nvidia A6000 GPU.

\subsection{Solution Constraints}
\label{sec:solution_contraints}

The proposed solution has few boundaries that affect its performance. First, the usage of GTM is less effective with high-contrast images compared to LTM methods. Second, consistency in bit-width between data simulation and inference is necessary, requiring conversion or re-training if different bit-widths are used. Third, a dedicated image histogram calculation module is required in the ISP.

\section{Experiments}
\label{sec:experiments}

The model was trained only with synthetically simulated data, but the testing was done both with synthetic and real data. The training dataset comprises 8-bit sRGB images, whereas our target input data comprises 26-bit unmapped images. However, given the use of 256-bin histograms, an 8-bit image is just enough to generate the required input. Consequently, using a higher bit-width image with the same number of histogram bins would yield no difference. In the end, the pipelines are designed and tested to work consistently across various bit-widths.

The results with real images have been processed according to Figure~\ref{fig:inference_pipeline}. However, for simulated images, the process involved a combination of the data simulation pipeline and the inference pipeline. In simulated results, comparisons were made between predictions and the original ground truth, utilizing both peak signal-to-noise ratio (PSNR) and visual evaluation metrics. Conversely, for real images, no ground truth was available, so comparisons were made against results from manual tuning of the tone curve.

\subsection{Histogram Evaluation}

Using a linear image histogram is effective for daylight images but becomes highly inaccurate in darker scenes. In a 26-bit image with a 256-bin histogram, the first bin accounts for pixel values ranging from 0 to $2^{18}-1 = 262,143$. This wide range makes it impossible to accurately estimate the required gain for the tone curve. For instance, in some tested midnight videos, the median unmapped linear image value was 30, requiring a vastly different gain compared to an image with a median of 100,000. Despite this, both scenarios would yield similar-looking linear histograms, resulting in inaccurate tone curve predictions.

In dark scenes, the vast majority of pixel values are concentrated in the first bin of the linear histogram, posing a challenge in predicting whether the appropriate gain should be 100 or 20,000. As illustrated in Figure~\ref{fig:applying_inverted_curve}, the linear histogram predominantly contains values within the first three bins when mapping with an inverted tone curve using a gain of 100. Hence, the logarithmic histogram proves necessary when dealing with a wide range of dark scenes. However, relying solely on a logarithmic histogram did not yield satisfactory performance, despite often providing a broader range of value distributions.

\subsection{Results with Simulated Images}
\label{sec:results_with_simulated_images}

\newlength{\simulatedresultswidth}
\setlength{\simulatedresultswidth}{0.18\columnwidth}
\ifdefined\ifConferenceTemplate\else
    \setlength{\simulatedresultswidth}{0.30\linewidth}
\fi

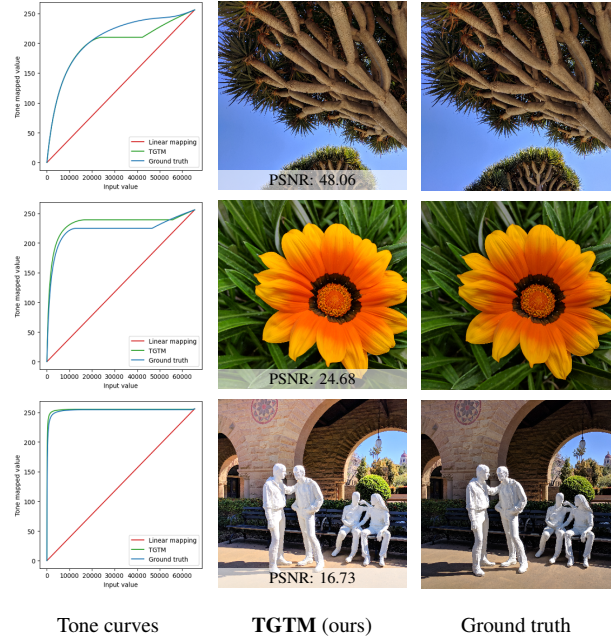
\begin{figure}
    \centering
    % 1st row
    \begin{subfigure}{\simulatedresultswidth}
        \begin{tikzpicture}
            \node[anchor=south west,inner sep=0] (image) at (0,0) 
            {\includegraphics[width=\linewidth]{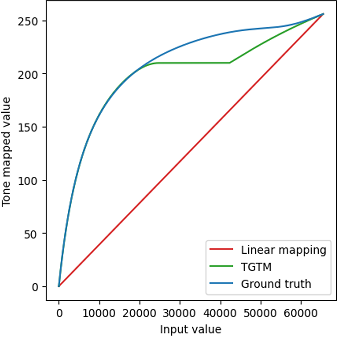}};
        \end{tikzpicture}
    \end{subfigure}
    %\hspace{1pt}
    %\begin{subfigure}{\simulatedresultswidth}
    %    \begin{tikzpicture}
    %        \node[anchor=south west,inner sep=0] (image) at (0,0) 
    %        {\includegraphics[width=\linewidth]{images/simulated_results/linear_histogram/1.png}};
    %    \end{tikzpicture}
    %\end{subfigure}
    \hspace{1pt}
    \begin{subfigure}{\simulatedresultswidth}
        \begin{tikzpicture}
            \node[anchor=south west,inner sep=0] (image) at (0,0) 
            {\includegraphics[width=\linewidth]{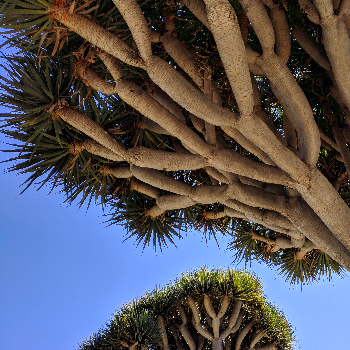}};
            \node[scale=0.6,fill=white,fill opacity=0.6,anchor=south,text width=\linewidth+0pt,
            inner xsep=24pt,inner ysep=3pt,outer sep=0pt,align=center,text opacity=1] at (image.south){PSNR: 48.06};
        \end{tikzpicture}
    \end{subfigure}
    \hspace{1pt}
    \begin{subfigure}{\simulatedresultswidth}
        \begin{tikzpicture}
            \node[anchor=south west,inner sep=0] (image) at (0,0) 
            {\includegraphics[width=\linewidth]{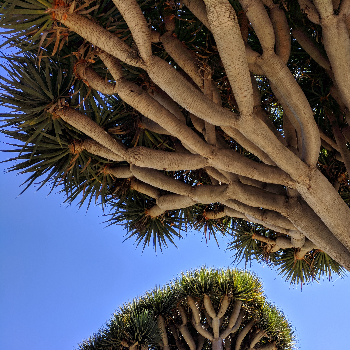}};
        \end{tikzpicture}
    \end{subfigure}
    \par\smallbreak
    % 2nd row
    \begin{subfigure}{\simulatedresultswidth}
        \begin{tikzpicture}
            \node[anchor=south west,inner sep=0] (image) at (0,0) 
            {\includegraphics[width=\linewidth]{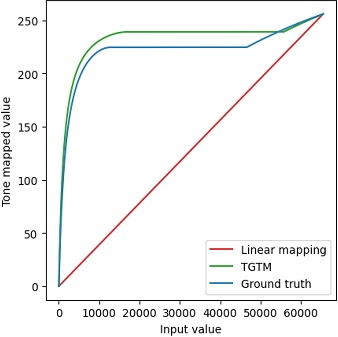}};
        \end{tikzpicture}
    \end{subfigure}
    %\hspace{1pt}
    %\begin{subfigure}{\simulatedresultswidth}
    %    \begin{tikzpicture}
    %        \node[anchor=south west,inner sep=0] (image) at (0,0) 
    %        {\includegraphics[width=\linewidth]{images/simulated_results/linear_histogram/39.png}};
    %    \end{tikzpicture}
    %\end{subfigure}
    \hspace{1pt}
    \begin{subfigure}{\simulatedresultswidth}
        \begin{tikzpicture}
            \node[anchor=south west,inner sep=0] (image) at (0,0) 
            {\includegraphics[width=\linewidth]{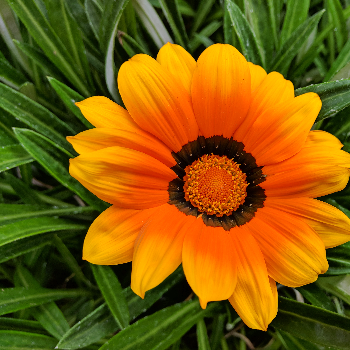}};
            \node[scale=0.6,fill=white,fill opacity=0.6,anchor=south,text width=\linewidth+0pt,
            inner xsep=24pt,inner ysep=3pt,outer sep=0pt,align=center,text opacity=1] at (image.south){PSNR: 24.68};
        \end{tikzpicture}
    \end{subfigure}
    \hspace{1pt}
    \begin{subfigure}{\simulatedresultswidth}
        \begin{tikzpicture}
            \node[anchor=south west,inner sep=0] (image) at (0,0) 
            {\includegraphics[width=\linewidth]{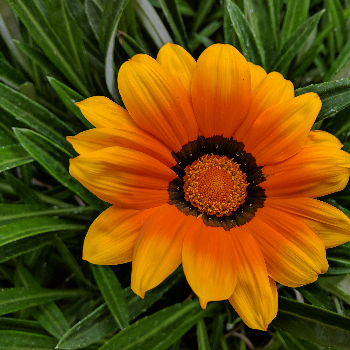}};
        \end{tikzpicture}
    \end{subfigure}
    \par\smallbreak
    % 3rd row
    \begin{subfigure}{\simulatedresultswidth}
        \begin{tikzpicture}
            \node[anchor=south west,inner sep=0] (image) at (0,0) 
            {\includegraphics[width=\linewidth]{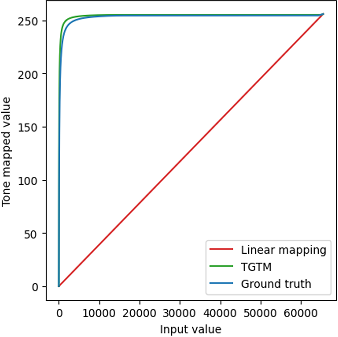}};
        \end{tikzpicture}
    \end{subfigure}
    %\hspace{1pt}
    %\begin{subfigure}{\simulatedresultswidth}
    %    \begin{tikzpicture}
    %        \node[anchor=south west,inner sep=0] (image) at (0,0) 
    %        {\includegraphics[width=\linewidth]{images/simulated_results/linear_histogram/18.png}};
    %    \end{tikzpicture}
    %\end{subfigure}
    \hspace{1pt}
    \begin{subfigure}{\simulatedresultswidth}
        \begin{tikzpicture}
            \node[anchor=south west,inner sep=0] (image) at (0,0) 
            {\includegraphics[width=\linewidth]{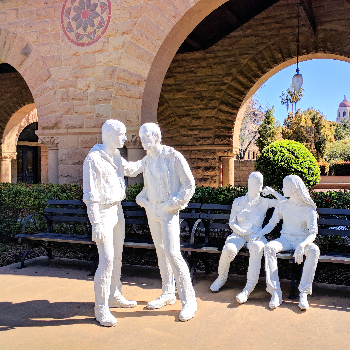}};
            \node[scale=0.6,fill=white,fill opacity=0.6,anchor=south,text width=\linewidth+0pt,
            inner xsep=24pt,inner ysep=3pt,outer sep=0pt,align=center,text opacity=1] at (image.south){PSNR: 16.73};
        \end{tikzpicture}
    \end{subfigure}
    \hspace{1pt}
    \begin{subfigure}{\simulatedresultswidth}
        \centering
        \begin{tikzpicture}
            \node[anchor=south west,inner sep=0] (image) at (0,0) 
            {\includegraphics[width=\linewidth]{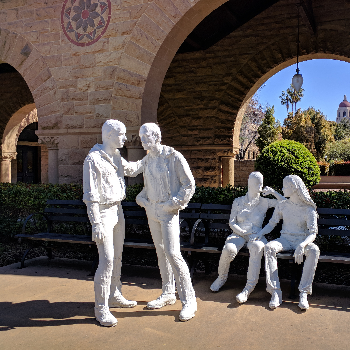}};
        \end{tikzpicture}
    \end{subfigure}
    \par\smallbreak
    \begin{subfigure}{\simulatedresultswidth}
        \centering
        \caption*{Tone curves}
    \end{subfigure}
    %\hspace{1pt}
    %\begin{subfigure}{\simulatedresultswidth}
    %    \centering
    %    \captionsetup{justification=centering}
    %    \caption*{Linear \\ histogram}
    %\end{subfigure}
    \hspace{1pt}
    \begin{subfigure}{\simulatedresultswidth}
        \caption*{\textbf{TGTM} (ours)}
    \end{subfigure}
    \hspace{1pt}
    \begin{subfigure}{\simulatedresultswidth}
        \centering
        \caption*{Ground truth}
    \end{subfigure}
    \caption{Three examples where PSNR of the TGTM outputs are higher, similar or lower than the average of 25.41~dB.}
    \label{fig:simulated_results}
\end{figure}

The performance was assessed using 100 samples from HDR+ dataset by simulating (Figure~\ref{fig:data_simulation_pipeline}) images and comparing the predictions (Figure~\ref{fig:inference_pipeline}) to the ground truth. Note that these images were not part of the training or validation datasets. On average, the prediction PSNR reached 25.41 dB with a standard deviation of 9.55. For a visual representation of these results, see Figure~\ref{fig:simulated_results}.

To the best of our knowledge, comprehensive benchmark datasets for global tone mapping of HDR sensors have not been established. While we could have compared our method with other methods using our data simulation pipeline, doing so would have unfairly favored our proposed method. Therefore, to maintain transparency and prevent any potential misinterpretation, we opted not to include simulated image comparisons with other methods in this publication.

\subsection{Results with Real Images}

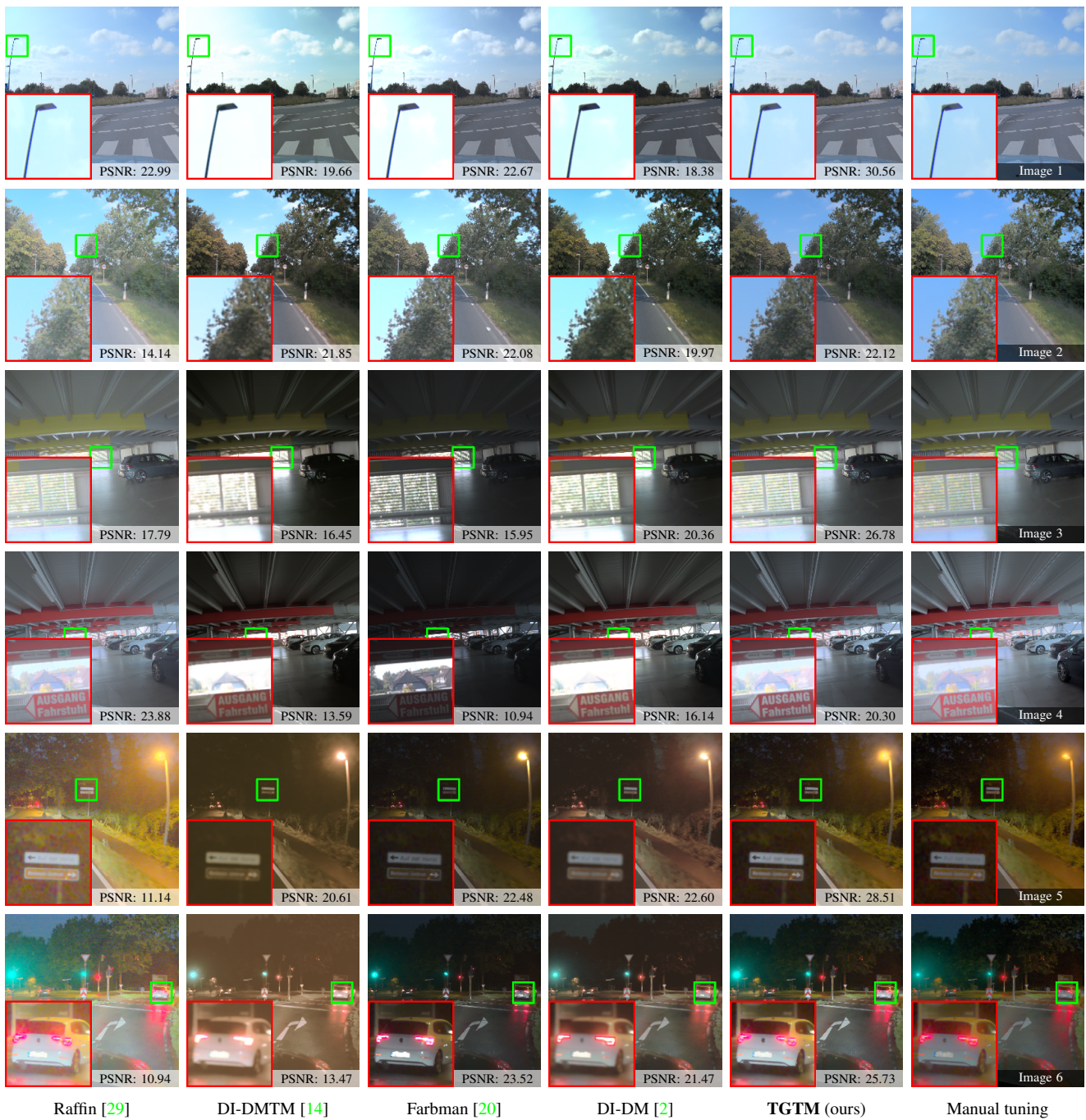
\begin{figure*}[!ht]
    \centering
    % 1st row
    \begin{subfigure}{0.16\linewidth}
        \centering
        \begin{tikzpicture}
            \node[anchor=south west,inner sep=0] (image) at (0,0) 
            {\includegraphics[width=\linewidth]{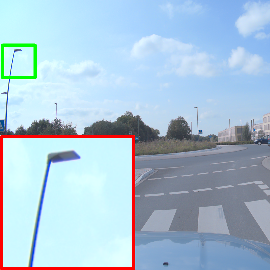}};
            \node[scale=0.6,fill=white,fill opacity=0.6,anchor=south west,text width=\linewidth-19pt,
            inner xsep=3pt,inner ysep=3pt,outer sep=0pt,align=center,text opacity=1] at (image.south){PSNR: 22.99};
            % from https://tex.stackexchange.com/a/9562/121799
        \end{tikzpicture}
    \end{subfigure}
    \hfill
    \begin{subfigure}{0.16\linewidth}
        \centering
        \begin{tikzpicture}
            \node[anchor=south west,inner sep=0] (image) at (0,0) 
            {\includegraphics[width=\linewidth]{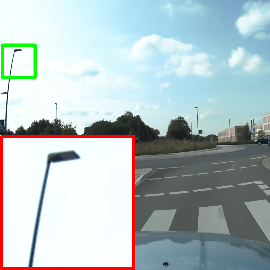}};
            \node[scale=0.6,fill=white,fill opacity=0.6,anchor=south west,text width=\linewidth-19pt,
            inner xsep=3pt,inner ysep=3pt,outer sep=0pt,align=center,text opacity=1] at (image.south){PSNR: 19.66};
            % from https://tex.stackexchange.com/a/9562/121799
        \end{tikzpicture}
    \end{subfigure}
    \hfill
    \begin{subfigure}{0.16\linewidth}
        \centering
        \begin{tikzpicture}
            \node[anchor=south west,inner sep=0] (image) at (0,0) 
            {\includegraphics[width=\linewidth]{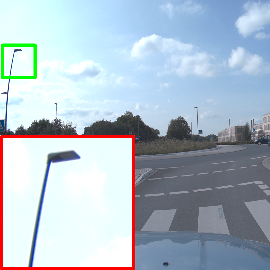}};
            \node[scale=0.6,fill=white,fill opacity=0.6,anchor=south west,text width=\linewidth-19pt,
            inner xsep=3pt,inner ysep=3pt,outer sep=0pt,align=center,text opacity=1] at (image.south){PSNR: 22.67};
            % from https://tex.stackexchange.com/a/9562/121799
        \end{tikzpicture}
    \end{subfigure}
    \hfill
    \begin{subfigure}{0.16\linewidth}
        \centering
        \begin{tikzpicture}
            \node[anchor=south west,inner sep=0] (image) at (0,0) 
            {\includegraphics[width=\linewidth]{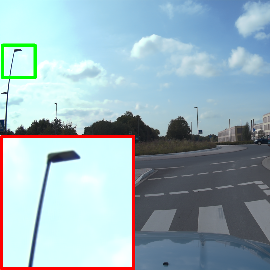}};
            \node[scale=0.6,fill=white,fill opacity=0.6,anchor=south west,text width=\linewidth-19pt,
            inner xsep=3pt,inner ysep=3pt,outer sep=0pt,align=center,text opacity=1] at (image.south){PSNR: 18.38};
            % from https://tex.stackexchange.com/a/9562/121799
        \end{tikzpicture}
    \end{subfigure}
    \hfill
    \begin{subfigure}{0.16\linewidth}
        \centering
        \begin{tikzpicture}
            \node[anchor=south west,inner sep=0] (image) at (0,0) 
            {\includegraphics[width=\linewidth]{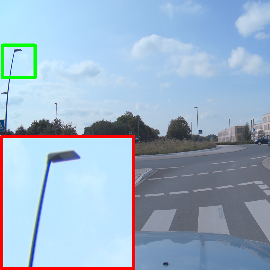}};
            \node[scale=0.6,fill=white,fill opacity=0.6,anchor=south west,text width=\linewidth-19pt,
            inner xsep=3pt,inner ysep=3pt,outer sep=0pt,align=center,text opacity=1] at (image.south){PSNR: 30.56};
            % from https://tex.stackexchange.com/a/9562/121799
        \end{tikzpicture}
    \end{subfigure}
    \hfill
    \begin{subfigure}{0.16\linewidth}
        \centering
        \begin{tikzpicture}
            \node[anchor=south west,inner sep=0] (image) at (0,0) 
            {\includegraphics[width=\linewidth]{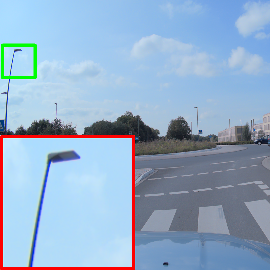}};
            \node[scale=0.6,fill=black,fill opacity=0.6,anchor=south west,text width=\linewidth-19pt,
            inner xsep=3pt,inner ysep=2pt,outer sep=0pt,align=center,text opacity=1] at (image.south){\textcolor{white}{Image 1}};
            % from https://tex.stackexchange.com/a/9562/121799
        \end{tikzpicture}
    \end{subfigure}
    \par\smallskip
    % 2nd row
    \begin{subfigure}{0.16\linewidth}
        \centering
        \begin{tikzpicture}
            \node[anchor=south west,inner sep=0] (image) at (0,0) 
            {\includegraphics[width=\linewidth]{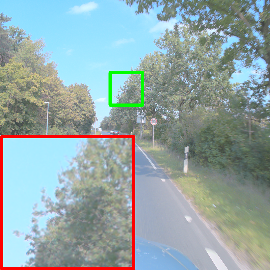}};
            \node[scale=0.6,fill=white,fill opacity=0.6,anchor=south west,text width=\linewidth-19pt,
            inner xsep=3pt,inner ysep=3pt,outer sep=0pt,align=center,text opacity=1] at (image.south){PSNR: 14.14};
            % from https://tex.stackexchange.com/a/9562/121799
        \end{tikzpicture}
    \end{subfigure}
    \hfill
    \begin{subfigure}{0.16\linewidth}
        \centering
        \begin{tikzpicture}
            \node[anchor=south west,inner sep=0] (image) at (0,0) 
            {\includegraphics[width=\linewidth]{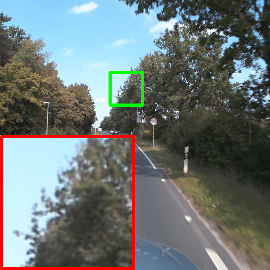}};
            \node[scale=0.6,fill=white,fill opacity=0.6,anchor=south west,text width=\linewidth-19pt,
            inner xsep=3pt,inner ysep=3pt,outer sep=0pt,align=center,text opacity=1] at (image.south){PSNR: 21.85};
            % from https://tex.stackexchange.com/a/9562/121799
        \end{tikzpicture}
    \end{subfigure}
    \hfill
    \begin{subfigure}{0.16\linewidth}
        \centering
        \begin{tikzpicture}
            \node[anchor=south west,inner sep=0] (image) at (0,0) 
            {\includegraphics[width=\linewidth]{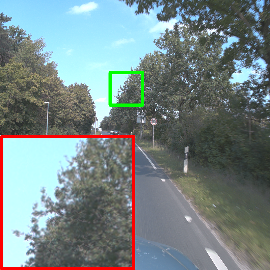}};
            \node[scale=0.6,fill=white,fill opacity=0.6,anchor=south west,text width=\linewidth-19pt,
            inner xsep=3pt,inner ysep=3pt,outer sep=0pt,align=center,text opacity=1] at (image.south){PSNR: 22.08};
            % from https://tex.stackexchange.com/a/9562/121799
        \end{tikzpicture}
    \end{subfigure}
    \hfill
    \begin{subfigure}{0.16\linewidth}
        \centering
        \begin{tikzpicture}
            \node[anchor=south west,inner sep=0] (image) at (0,0) 
            {\includegraphics[width=\linewidth]{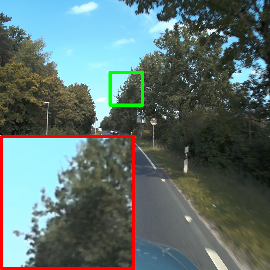}};
            \node[scale=0.6,fill=white,fill opacity=0.6,anchor=south west,text width=\linewidth-19pt,
            inner xsep=3pt,inner ysep=3pt,outer sep=0pt,align=center,text opacity=1] at (image.south){PSNR: 19.97};
            % from https://tex.stackexchange.com/a/9562/121799
        \end{tikzpicture}
    \end{subfigure}
    \hfill
    \begin{subfigure}{0.16\linewidth}
        \centering
        \begin{tikzpicture}
            \node[anchor=south west,inner sep=0] (image) at (0,0) 
            {\includegraphics[width=\linewidth]{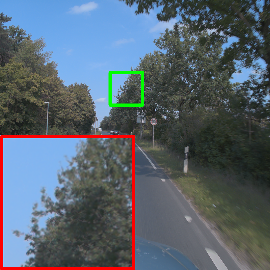}};
            \node[scale=0.6,fill=white,fill opacity=0.6,anchor=south west,text width=\linewidth-19pt,
            inner xsep=3pt,inner ysep=3pt,outer sep=0pt,align=center,text opacity=1] at (image.south){PSNR: 22.12};
            % from https://tex.stackexchange.com/a/9562/121799
        \end{tikzpicture}
    \end{subfigure}
    \hfill
    \begin{subfigure}{0.16\linewidth}
        \centering
        \begin{tikzpicture}
            \node[anchor=south west,inner sep=0] (image) at (0,0) 
            {\includegraphics[width=\linewidth]{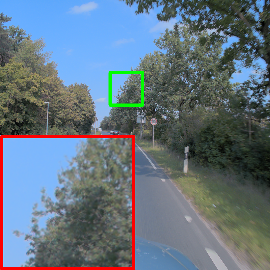}};
            \node[scale=0.6,fill=black,fill opacity=0.6,anchor=south west,text width=\linewidth-19pt,
            inner xsep=3pt,inner ysep=2pt,outer sep=0pt,align=center,text opacity=1] at (image.south){\textcolor{white}{Image 2}};
            % from https://tex.stackexchange.com/a/9562/121799
        \end{tikzpicture}
    \end{subfigure}
    \par\smallskip
    % 3rd row
    \begin{subfigure}{0.16\linewidth}
        \centering
        \begin{tikzpicture}
            \node[anchor=south west,inner sep=0] (image) at (0,0) 
            {\includegraphics[width=\linewidth]{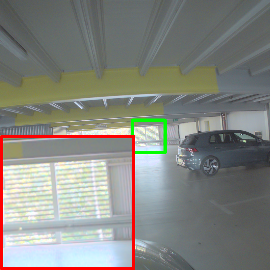}};
            \node[scale=0.6,fill=white,fill opacity=0.6,anchor=south west,text width=\linewidth-19pt,
            inner xsep=3pt,inner ysep=3pt,outer sep=0pt,align=center,text opacity=1] at (image.south){PSNR: 17.79};
            % from https://tex.stackexchange.com/a/9562/121799
        \end{tikzpicture}
    \end{subfigure}
    \hfill
    \begin{subfigure}{0.16\linewidth}
        \centering
        \begin{tikzpicture}
            \node[anchor=south west,inner sep=0] (image) at (0,0) 
            {\includegraphics[width=\linewidth]{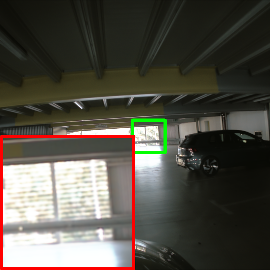}};
            \node[scale=0.6,fill=white,fill opacity=0.6,anchor=south west,text width=\linewidth-19pt,
            inner xsep=3pt,inner ysep=3pt,outer sep=0pt,align=center,text opacity=1] at (image.south){PSNR: 16.45};
            % from https://tex.stackexchange.com/a/9562/121799
        \end{tikzpicture}
    \end{subfigure}
    \hfill
    \begin{subfigure}{0.16\linewidth}
        \centering
        \begin{tikzpicture}
            \node[anchor=south west,inner sep=0] (image) at (0,0) 
            {\includegraphics[width=\linewidth]{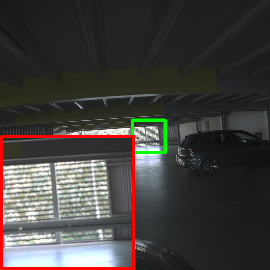}};
            \node[scale=0.6,fill=white,fill opacity=0.6,anchor=south west,text width=\linewidth-19pt,
            inner xsep=3pt,inner ysep=3pt,outer sep=0pt,align=center,text opacity=1] at (image.south){PSNR: 15.95};
            % from https://tex.stackexchange.com/a/9562/121799
        \end{tikzpicture}
    \end{subfigure}
    \hfill
    \begin{subfigure}{0.16\linewidth}
        \centering
        \begin{tikzpicture}
            \node[anchor=south west,inner sep=0] (image) at (0,0) 
            {\includegraphics[width=\linewidth]{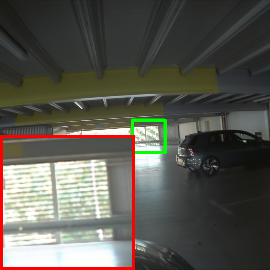}};
            \node[scale=0.6,fill=white,fill opacity=0.6,anchor=south west,text width=\linewidth-19pt,
            inner xsep=3pt,inner ysep=3pt,outer sep=0pt,align=center,text opacity=1] at (image.south){PSNR: 20.36};
            % from https://tex.stackexchange.com/a/9562/121799
        \end{tikzpicture}
    \end{subfigure}
    \hfill
    \begin{subfigure}{0.16\linewidth}
        \centering
        \begin{tikzpicture}
            \node[anchor=south west,inner sep=0] (image) at (0,0) 
            {\includegraphics[width=\linewidth]{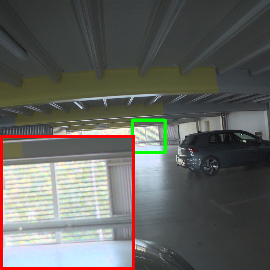}};
            \node[scale=0.6,fill=white,fill opacity=0.6,anchor=south west,text width=\linewidth-19pt,
            inner xsep=3pt,inner ysep=3pt,outer sep=0pt,align=center,text opacity=1] at (image.south){PSNR: 26.78};
            % from https://tex.stackexchange.com/a/9562/121799
        \end{tikzpicture}
    \end{subfigure}
    \hfill
    \begin{subfigure}{0.16\linewidth}
        \centering
        \begin{tikzpicture}
            \node[anchor=south west,inner sep=0] (image) at (0,0) 
            {\includegraphics[width=\linewidth]{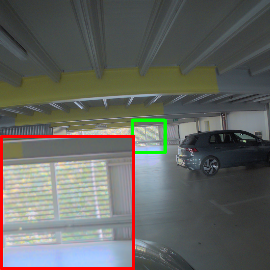}};
            \node[scale=0.6,fill=black,fill opacity=0.6,anchor=south west,text width=\linewidth-19pt,
            inner xsep=3pt,inner ysep=2pt,outer sep=0pt,align=center,text opacity=1] at (image.south){\textcolor{white}{Image 3}};
            % from https://tex.stackexchange.com/a/9562/121799
        \end{tikzpicture}
    \end{subfigure}
    \par\smallbreak
    % 4th row
    \begin{subfigure}{0.16\linewidth}
        \centering
        \begin{tikzpicture}
            \node[anchor=south west,inner sep=0] (image) at (0,0) 
            {\includegraphics[width=\linewidth]{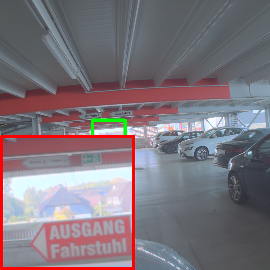}};
            \node[scale=0.6,fill=white,fill opacity=0.6,anchor=south west,text width=\linewidth-19pt,
            inner xsep=3pt,inner ysep=3pt,outer sep=0pt,align=center,text opacity=1] at (image.south){PSNR: 23.88};
            % from https://tex.stackexchange.com/a/9562/121799
        \end{tikzpicture}
    \end{subfigure}
    \hfill
    \begin{subfigure}{0.16\linewidth}
        \centering
        \begin{tikzpicture}
            \node[anchor=south west,inner sep=0] (image) at (0,0) 
            {\includegraphics[width=\linewidth]{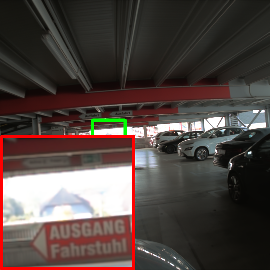}};
            \node[scale=0.6,fill=white,fill opacity=0.6,anchor=south west,text width=\linewidth-19pt,
            inner xsep=3pt,inner ysep=3pt,outer sep=0pt,align=center,text opacity=1] at (image.south){PSNR: 13.59};
            % from https://tex.stackexchange.com/a/9562/121799
        \end{tikzpicture}
    \end{subfigure}
    \hfill
    \begin{subfigure}{0.16\linewidth}
        \centering
        \begin{tikzpicture}
            \node[anchor=south west,inner sep=0] (image) at (0,0) 
            {\includegraphics[width=\linewidth]{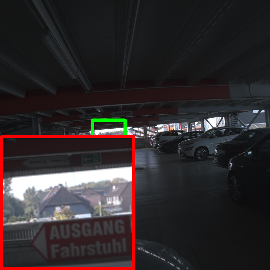}};
            \node[scale=0.6,fill=white,fill opacity=0.6,anchor=south west,text width=\linewidth-19pt,
            inner xsep=3pt,inner ysep=3pt,outer sep=0pt,align=center,text opacity=1] at (image.south){PSNR: 10.94};
            % from https://tex.stackexchange.com/a/9562/121799
        \end{tikzpicture}
    \end{subfigure}
    \hfill
    \begin{subfigure}{0.16\linewidth}
        \centering
        \begin{tikzpicture}
            \node[anchor=south west,inner sep=0] (image) at (0,0) 
            {\includegraphics[width=\linewidth]{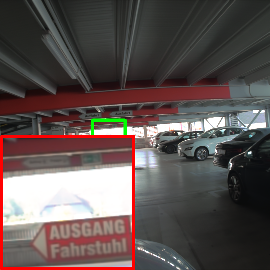}};
            \node[scale=0.6,fill=white,fill opacity=0.6,anchor=south west,text width=\linewidth-19pt,
            inner xsep=3pt,inner ysep=3pt,outer sep=0pt,align=center,text opacity=1] at (image.south){PSNR: 16.14};
            % from https://tex.stackexchange.com/a/9562/121799
        \end{tikzpicture}
    \end{subfigure}
    \hfill
    \begin{subfigure}{0.16\linewidth}
        \centering
        \begin{tikzpicture}
            \node[anchor=south west,inner sep=0] (image) at (0,0) 
            {\includegraphics[width=\linewidth]{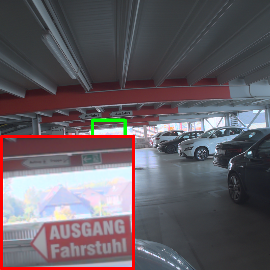}};
            \node[scale=0.6,fill=white,fill opacity=0.6,anchor=south west,text width=\linewidth-19pt,
            inner xsep=3pt,inner ysep=3pt,outer sep=0pt,align=center,text opacity=1] at (image.south){PSNR: 20.30};
            % from https://tex.stackexchange.com/a/9562/121799
        \end{tikzpicture}
    \end{subfigure}
    \hfill
    \begin{subfigure}{0.16\linewidth}
        \centering
        \begin{tikzpicture}
            \node[anchor=south west,inner sep=0] (image) at (0,0) 
            {\includegraphics[width=\linewidth]{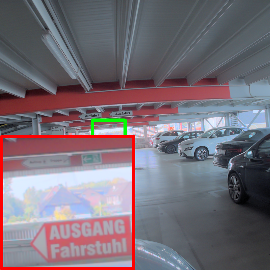}};
            \node[scale=0.6,fill=black,fill opacity=0.6,anchor=south west,text width=\linewidth-19pt,
            inner xsep=3pt,inner ysep=2pt,outer sep=0pt,align=center,text opacity=1] at (image.south){\textcolor{white}{Image 4}};
            % from https://tex.stackexchange.com/a/9562/121799
        \end{tikzpicture}
    \end{subfigure}
    \par\smallskip
    % 5th row
    \begin{subfigure}{0.16\linewidth}
        \centering
        \begin{tikzpicture}
            \node[anchor=south west,inner sep=0] (image) at (0,0) 
            {\includegraphics[width=\linewidth]{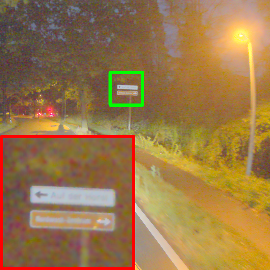}};
            \node[scale=0.6,fill=white,fill opacity=0.6,anchor=south west,text width=\linewidth-19pt,
            inner xsep=3pt,inner ysep=3pt,outer sep=0pt,align=center,text opacity=1] at (image.south){PSNR: 11.14};
            % from https://tex.stackexchange.com/a/9562/121799
        \end{tikzpicture}
    \end{subfigure}
    \hfill
    \begin{subfigure}{0.16\linewidth}
        \centering
        \begin{tikzpicture}
            \node[anchor=south west,inner sep=0] (image) at (0,0) 
            {\includegraphics[width=\linewidth]{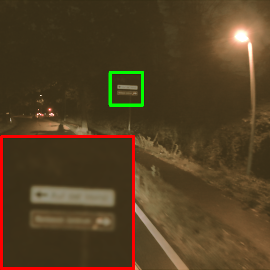}};
            \node[scale=0.6,fill=white,fill opacity=0.6,anchor=south west,text width=\linewidth-19pt,
            inner xsep=3pt,inner ysep=3pt,outer sep=0pt,align=center,text opacity=1] at (image.south){PSNR: 20.61};
            % from https://tex.stackexchange.com/a/9562/121799
        \end{tikzpicture}
    \end{subfigure}
    \hfill
    \begin{subfigure}{0.16\linewidth}
        \centering
        \begin{tikzpicture}
            \node[anchor=south west,inner sep=0] (image) at (0,0) 
            {\includegraphics[width=\linewidth]{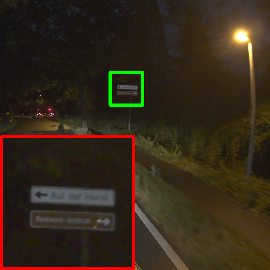}};
            \node[scale=0.6,fill=white,fill opacity=0.6,anchor=south west,text width=\linewidth-19pt,
            inner xsep=3pt,inner ysep=3pt,outer sep=0pt,align=center,text opacity=1] at (image.south){PSNR: 22.48};
            % from https://tex.stackexchange.com/a/9562/121799
        \end{tikzpicture}
    \end{subfigure}
    \hfill
    \begin{subfigure}{0.16\linewidth}
        \centering
        \begin{tikzpicture}
            \node[anchor=south west,inner sep=0] (image) at (0,0) 
            {\includegraphics[width=\linewidth]{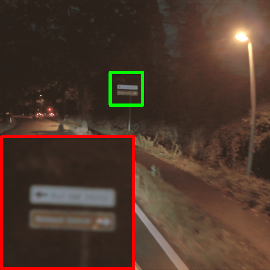}};
            \node[scale=0.6,fill=white,fill opacity=0.6,anchor=south west,text width=\linewidth-19pt,
            inner xsep=3pt,inner ysep=3pt,outer sep=0pt,align=center,text opacity=1] at (image.south){PSNR: 22.60};
            % from https://tex.stackexchange.com/a/9562/121799
        \end{tikzpicture}
    \end{subfigure}
    \hfill
    \begin{subfigure}{0.16\linewidth}
        \centering
        \begin{tikzpicture}
            \node[anchor=south west,inner sep=0] (image) at (0,0) 
            {\includegraphics[width=\linewidth]{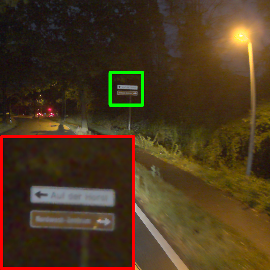}};
            \node[scale=0.6,fill=white,fill opacity=0.6,anchor=south west,text width=\linewidth-19pt,
            inner xsep=3pt,inner ysep=3pt,outer sep=0pt,align=center,text opacity=1] at (image.south){PSNR: 28.51};
            % from https://tex.stackexchange.com/a/9562/121799
        \end{tikzpicture}
    \end{subfigure}
    \hfill
    \begin{subfigure}{0.16\linewidth}
        \centering
        \begin{tikzpicture}
            \node[anchor=south west,inner sep=0] (image) at (0,0) 
            {\includegraphics[width=\linewidth]{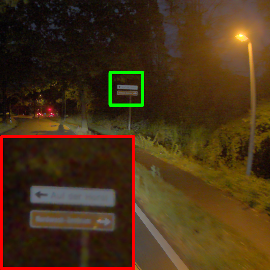}};
            \node[scale=0.6,fill=black,fill opacity=0.6,anchor=south west,text width=\linewidth-19pt,
            inner xsep=3pt,inner ysep=2pt,outer sep=0pt,align=center,text opacity=1] at (image.south){\textcolor{white}{Image 5}};
            % from https://tex.stackexchange.com/a/9562/121799
        \end{tikzpicture}
    \end{subfigure}
    \par\smallskip
    % 6th row
    \begin{subfigure}{0.16\linewidth}
        \centering
        \begin{tikzpicture}
            \node[anchor=south west,inner sep=0] (image) at (0,0) 
            {\includegraphics[width=\linewidth]{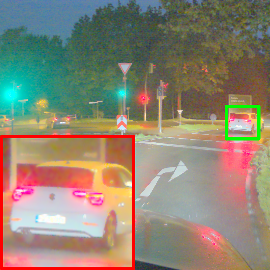}};
            \node[scale=0.6,fill=white,fill opacity=0.6,anchor=south west,text width=\linewidth-19pt,
            inner xsep=3pt,inner ysep=3pt,outer sep=0pt,align=center,text opacity=1] at (image.south){PSNR: 10.94};
            % from https://tex.stackexchange.com/a/9562/121799
        \end{tikzpicture}
        \caption*{Raffin~\cite{raffin}}
    \end{subfigure}
    \hfill
    \begin{subfigure}{0.16\linewidth}
        \centering
        \begin{tikzpicture}
            \node[anchor=south west,inner sep=0] (image) at (0,0) 
            {\includegraphics[width=\linewidth]{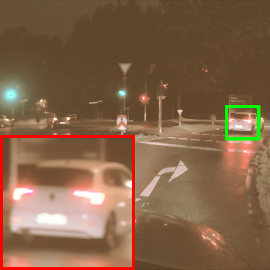}};
            \node[scale=0.6,fill=white,fill opacity=0.6,anchor=south west,text width=\linewidth-19pt,
            inner xsep=3pt,inner ysep=3pt,outer sep=0pt,align=center,text opacity=1] at (image.south){PSNR: 13.47};
            % from https://tex.stackexchange.com/a/9562/121799
        \end{tikzpicture}
        \caption*{DI-DMTM~\cite{hdr_stojkovic}}
    \end{subfigure}
    \hfill
    \begin{subfigure}{0.16\linewidth}
        \centering
        \begin{tikzpicture}
            \node[anchor=south west,inner sep=0] (image) at (0,0) 
            {\includegraphics[width=\linewidth]{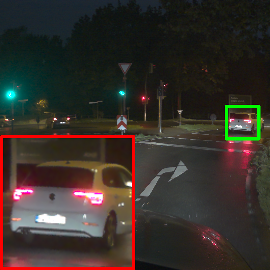}};
            \node[scale=0.6,fill=white,fill opacity=0.6,anchor=south west,text width=\linewidth-19pt,
            inner xsep=3pt,inner ysep=3pt,outer sep=0pt,align=center,text opacity=1] at (image.south){PSNR: 23.52};
            % from https://tex.stackexchange.com/a/9562/121799
        \end{tikzpicture}
        \caption*{Farbman~\cite{farbman_2008_edge}}
    \end{subfigure}
    \hfill
    \begin{subfigure}{0.16\linewidth}
        \centering
        \begin{tikzpicture}
            \node[anchor=south west,inner sep=0] (image) at (0,0) 
            {\includegraphics[width=\linewidth]{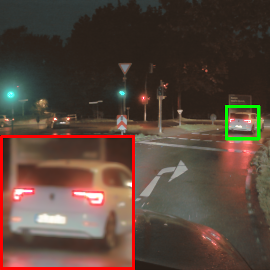}};
            \node[scale=0.6,fill=white,fill opacity=0.6,anchor=south west,text width=\linewidth-19pt,
            inner xsep=3pt,inner ysep=3pt,outer sep=0pt,align=center,text opacity=1] at (image.south){PSNR: 21.47};
            % from https://tex.stackexchange.com/a/9562/121799
        \end{tikzpicture}
        \caption*{DI-DM~\cite{hdr_shopovska}}
    \end{subfigure}
    \hfill
    \begin{subfigure}{0.16\linewidth}
        \centering
        \begin{tikzpicture}
            \node[anchor=south west,inner sep=0] (image) at (0,0) 
            {\includegraphics[width=\linewidth]{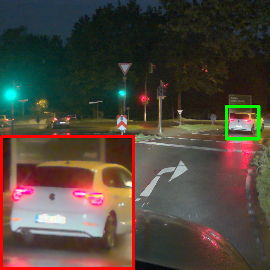}};
            \node[scale=0.6,fill=white,fill opacity=0.6,anchor=south west,text width=\linewidth-19pt,
            inner xsep=3pt,inner ysep=3pt,outer sep=0pt,align=center,text opacity=1] at (image.south){PSNR: 25.73};
            % from https://tex.stackexchange.com/a/9562/121799
        \end{tikzpicture}
        \caption*{\textbf{TGTM} (ours)}
    \end{subfigure}
    \hfill
    \begin{subfigure}{0.16\linewidth}
        \centering
        \begin{tikzpicture}
            \node[anchor=south west,inner sep=0] (image) at (0,0) 
            {\includegraphics[width=\linewidth]{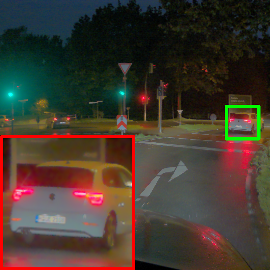}};
            \node[scale=0.6,fill=black,fill opacity=0.6,anchor=south west,text width=\linewidth-19pt,
            inner xsep=3pt,inner ysep=2pt,outer sep=0pt,align=center,text opacity=1] at (image.south){\textcolor{white}{Image 6}};
            % from https://tex.stackexchange.com/a/9562/121799
        \end{tikzpicture}
        \caption*{Manual tuning}
    \end{subfigure}
    \caption{Visual comparison of state-of-the-art methods using real-world 26-bit unmapped images. There are three different lighting conditions: daylight, combination of indoor and outdoor lights, and midnight light. The PSNR scores have been calculated between each method and manual tuning.}
    \label{fig:real_image_tone_mapping}
\end{figure*}

\begin{table*}[!ht]
    \caption{Quantitative comparison (PSNR) of state-of-the-art methods with the real 26-bit HDR camera images shown in Figure~\ref{fig:real_image_tone_mapping}. The best and the second best performance are \textbf{bolded} and \underline{underlined}, respectively.}
    \label{tab:real_image_tone_mapping}
    \vskip 0.15in
    \begin{center}
    \begin{small}
    \begin{rmfamily}
    \ifdefined\ifConferenceTemplate
        \begin{tabular}{lccccccc}
    \else
        \begin{tabular}{l|c|c|c|c|c|c|c}
    \fi
    \toprule
    Method & Image 1 & Image 2 & Image 3 & Image 4 & Image 5 & Image 6 & Average \\
    \midrule
    \textbf{TGTM} (ours) & \bf 30.56 & \bf 22.12 & \bf 26.78 & \underline{20.30} & \bf 28.51 & \bf 25.73 & \bf 25.67 \\
    DI-TM~\cite{hdr_shopovska} & 18.38 & 19.97 & \underline{20.36} & 16.14 & \underline{22.60} & 21.47 & \underline{19.82} \\
    Farbman~\cite{farbman_2008_edge} & 22.67 & \underline{22.08} & 15.95 & 10.94 & 22.48 & \underline{23.52} & 19.61 \\
    DI-DMTM~\cite{hdr_stojkovic} & 19.66 & 21.85 & 16.45 & 13.59 & 20.61 & 13.47 & 17.60 \\
    Raffin~\cite{raffin} & \underline{22.99} & 14.14 & 17.79 & \bf 23.88 & 11.14 & 10.94 & 16.81 \\
    \bottomrule
    \end{tabular}
    \end{rmfamily}
    \end{small}
    \end{center}
\end{table*}

The proposed tone mapping method was tested with multiple videos where we mounted an HDR camera (using an onsemi AR0341AT~\cite{onsemi_sensor} sensor) to a car. The dynamic range of the sensor is 150~dB and the image data is stored in 26-bit format. In our benchmark, we selected six 26-bit unmapped input images which all were tone mapped with the proposed algorithm and with four other algorithms. The results were compared to manual tuning that was independently performed in advance, see Figure~\ref{fig:real_image_tone_mapping}. The six images were selected with different conditions and challenging situations in mind. In most cases, our proposed solution was very close to the results of manual tuning, therefore we selected situations with the most difference to manual tuning. Subsequently, the six images were sent for processing of DI-TM, DI-DMTM, and Farbman algorithms to the authors of DI-TM. Raffin~\cite{raffin} results were processed by us. None of the algorithms, including ours, had prior knowledge of any of these images, ensuring a level playing field for all.

In the first image depicted in Figure~\ref{fig:real_image_tone_mapping}, all four comparison methods exhibit a cyan hue in the sky at the top left, while our proposed method, TGTM, produces the most natural-looking result for this scene. Moving to the second image, both DI-TM and DI-DMTM methods demonstrate reduction in details and intensified color saturation. Additionally, all four images in this set suffer from a cyan tint in the sky. However, TGTM stands out by preserving the natural and pleasing blue hue of the sky. In the bottom four images, DI-TM and DI-DMTM methods begin to exhibit clear smoothening artifacts. Particularly in images 3 and 4, DI-TM and DI-DMTM struggle to handle high-contrast situations, resulting in a darkened hall and an overly bright window. On the contrary, Farbman dims the hall further but enhances the quality of the window. Nevertheless, the resultant overall image remains excessively dark, falling short of achieving a visually pleasing outcome. Raffin effectively illuminates the hall to a degree surpassing the preference of our manual tuning, yet without excessively brightening the window, maintaining a satisfactory level of brightness. TGTM tone maps the halls slightly dimmer than manual tuning in images 3 and 4. Additionally in image 3, the window appears brighter compared to the result of manual tuning. On the contrary, in image 4, the window shows enhanced contrast in the sign. In the midnight scenes depicted in images 5 and 6, DI-TM and DI-DMTM exhibit blurriness. Additionally, DI-TM renders the trees excessively dark, while DI-DMTM introduces a noticeable yellowish tone to the image. Farbman maintains accurate colors but results in overly dark images, whereas Raffin does the opposite by brightening the image excessively. However, TGTM demonstrates a good balance in both brightness and contrast, particularly evident in the depiction of the street scene.

The quantitative results from Figure~\ref{fig:real_image_tone_mapping} have been combined to Table~\ref{tab:real_image_tone_mapping}. There we see TGTM outperforms on average the second best by 5.85~dB. The dataset is not large with the real HDR sensor but the three different lighting conditions are indicating the tone mapping performance of the algorithms in various scenes.

The general consensus regarding the CNN-based pixel-wise tone mapping methods suggests that DI-TM and DI-DMTM have a tendency to produce smoother results than manual tuning, often with added color aberrations. Additionally, blurry outcomes were obtained in the bottom four low-light images in Figure~\ref{fig:real_image_tone_mapping}. In all the six images, Raffin consistently produced brighter outcomes. However, in the top two daylight images, the tone of the sky was inaccurate. In contrast, Farbman had satisfactory performance in daylight images but yielded overly dark outcomes in dim and dark conditions. Our proposed method, TGTM, performed the best in these six images by 5.85 dB higher PSNR on average compared to the second best method. Overall TGTM performed closely with the results of manual tuning but TGTM could not handle the windows in the images 3 and 4 as well as manual tuning.

Temporal stability in video refers to the consistency and smoothness of motion over time. It indicates how well the video maintains its integrity, without jitter, flickering, or other disruptions that could distract the viewing experience. In addition to assessing single image quality, it is important to recognize that TGTM also exhibits robust temporal stability when evaluated across multiple videos processed frame by frame. Given our target application of delivering pleasing image quality to automotive settings, it is crucial to present temporally stable video content to avoid distracting drivers. Moreover, maintaining stable image quality is vital considering the machine vision decision making in ADAS.

\section{Conclusion}

In this paper, we propose a neural network-based global tone mapping pipeline suited for very high dynamic range mapping in an embedded system. The network takes in linear and logarithmic luminance histograms as inputs and predicts four parameters that are fed into an analytic tone curve function ensuring consistent curve shapes and temporal stability. By taking in 256-bin histograms, the input pixel data is highly compressed resulting in a tiny neural network that needs just 9 kFLOPS and only 1k model parameters. By applying inverted tone curves on already tone mapped images of a public dataset, a training dataset could be simulated, which proved sufficient for training our network. The low computational requirements of the proposed approach and its algorithmic stability at various lighting and contrast conditions makes it suitable for usage in an embedded image processing pipeline.

\vspace{0.5cm}
\normalsize{
	\noindent\textbf{Acknowledgment}.
	We would like to thank the authors of DI-TM for supporting us with processing the results of the DI-TM, DI-DMTM, and Farbman methods.\\
}
\vspace{-0.9cm}
%\clearpage
%\clearpage
{\small
\bibliographystyle{unsrt}
\bibliography{main}
}

\end{document}